%% file: main.tex
\documentclass[preprint,12pt,authoryear]{elsarticle}

\usepackage[table,xcdraw]{xcolor}
\usepackage{hyperref}
\usepackage{graphicx}
\usepackage{adjustbox}
\usepackage{multirow}
\usepackage{booktabs}
\usepackage{subfigure}
\usepackage{algorithm}
\usepackage{algorithmicx}
\usepackage[noend]{algpseudocode}
\usepackage{rotating}
\usepackage{amsthm}
\usepackage{amsmath}
\usepackage{amssymb}

\colorlet{red}{black}
\colorlet{blue}{black}

\journal{Medical Image Analysis}

\usepackage{lineno}

\begin{document}

\begin{frontmatter}



\title{Federated Modality-specific Encoders and Partially Personalized Fusion Decoder for Multimodal Brain Tumor Segmentation} 


\author[a,c]{Hong {Liu}\fnref{fn1}}\ead{liuhong@stu.xmu.edu.cn}
\author[c]{Dong {Wei}\fnref{fn1}}\ead{donwei@tencent.com}
\fntext[fn1]{Hong Liu and Dong Wei contributed equally. Hong Liu contributed to this work during an internship at Tencent.}
\author[b]{Qian {Dai}}\ead{daiqian@stu.xmu.edu.cn}
\author[c]{Xian {Wu}\corref{cor1}}\ead{kevinxwu@tencent.com}
\author[c,d]{Yefeng {Zheng}\corref{cor1}}\ead{yefengzheng@tencent.com}
\author[a,b]{Liansheng {Wang}\corref{cor1}}\ead{lswang@xmu.edu.cn}
\cortext[cor1]{Corresponding authors: L. Wang, X. Wu and Y. Zheng.}

\affiliation[a]{organization={National Institute for Data Science in Health and Medicine},
            addressline={Xiamen University}, 
            city={Xiamen},
            postcode={361005}, 
            state={Fujian},
            country={China}}
\affiliation[b]{organization={Department of Computer Science at School of Informatics},
            addressline={Xiamen University}, 
            city={Xiamen},
            postcode={361005}, 
            state={Fujian},
            country={China}}
\affiliation[c]{organization={Jarvis Research Center},
            addressline={Tencent YouTu Lab}, 
            city={Shenzhen},
            postcode={518075}, 
            state={Guangdong},
            country={China}}
\affiliation[d]{organization={Medical Artificial Intelligence Lab},
            addressline={Westlake University}, 
            city={Hangzhou},
            postcode={310030}, 
            state={Zhejiang},
            country={China}}

\begin{abstract}
Most existing federated learning (FL) methods for medical image analysis only considered intramodal heterogeneity, limiting their applicability to multimodal imaging applications.
In practice, some FL participants may possess only a subset of the complete imaging modalities, 
posing intermodal heterogeneity as a challenge to effectively training a global model on all participants' data.
Meanwhile, each participant expects a personalized model tailored to its local data characteristics in FL.
This work proposes a new FL framework with federated modality-specific encoders and partially personalized multimodal fusion decoders (FedMEPD) to address the two concurrent issues.
Specifically, FedMEPD employs an exclusive encoder for each modality to account for the intermodal heterogeneity.
{\color{blue}While these encoders are fully federated, the decoders are partially personalized to meet individual needs---using the discrepancy between global and local parameter updates to dynamically determine which decoder filters are personalized.}
Implementation-wise, a server with full-modal data employs a fusion decoder to fuse representations from all modality-specific encoders, thus bridging the modalities to optimize the encoders via backpropagation.
Moreover, multiple anchors are extracted from the fused multimodal representations and distributed to the clients in addition to the model parameters.
Conversely, the clients with incomplete modalities calibrate their missing-modal representations toward the global full-modal anchors via scaled dot-product cross-attention, making up for the information loss due to absent modalities.
FedMEPD is validated on the BraTS 2018 and 2020 multimodal brain tumor segmentation benchmarks.
Results show that it outperforms various up-to-date methods for multimodal and personalized FL, and its novel designs are effective.
\end{abstract}




\end{frontmatter}


\input{sections/1-introduction.tex}

\input{sections/related_work.tex}

\input{sections/2-method.tex}
\input{sections/3-experiments.tex}
\input{sections/4-discussion-conclusion.tex}

\section{Acknowledgments} This work was supported by the National Natural Science Foundation of China (Grant No. 62371409).




\bibliographystyle{elsarticle-harv}
\bibliography{references}

\end{document}

%% file: sections/1-introduction.tex
\section{Introduction}
\label{sec:intoduction}

Federated learning (FL) enables participants to collaboratively train a global model on their collective data without 
breaching privacy {\color{red}\citep{li2020federated,foley2022openfl,pati2022federated}}.
The decentralized mechanism makes it particularly suitable for privacy-sensitive applications such as medical image analysis~{\color{red}\citep{sheller2020federated,kaissis2020secure,adnan2022federated,yan2020variation,pati2021federated,pati2022federated,Pati_2022}}. 
However, most existing FL methods for medical image analysis only considered intramodal heterogeneity, limiting their applicability to multimodal imaging in practice.

\begin{figure}[t]
\centering\footnotesize\includegraphics[trim=0 0 0 0, clip, width=.9\textwidth]{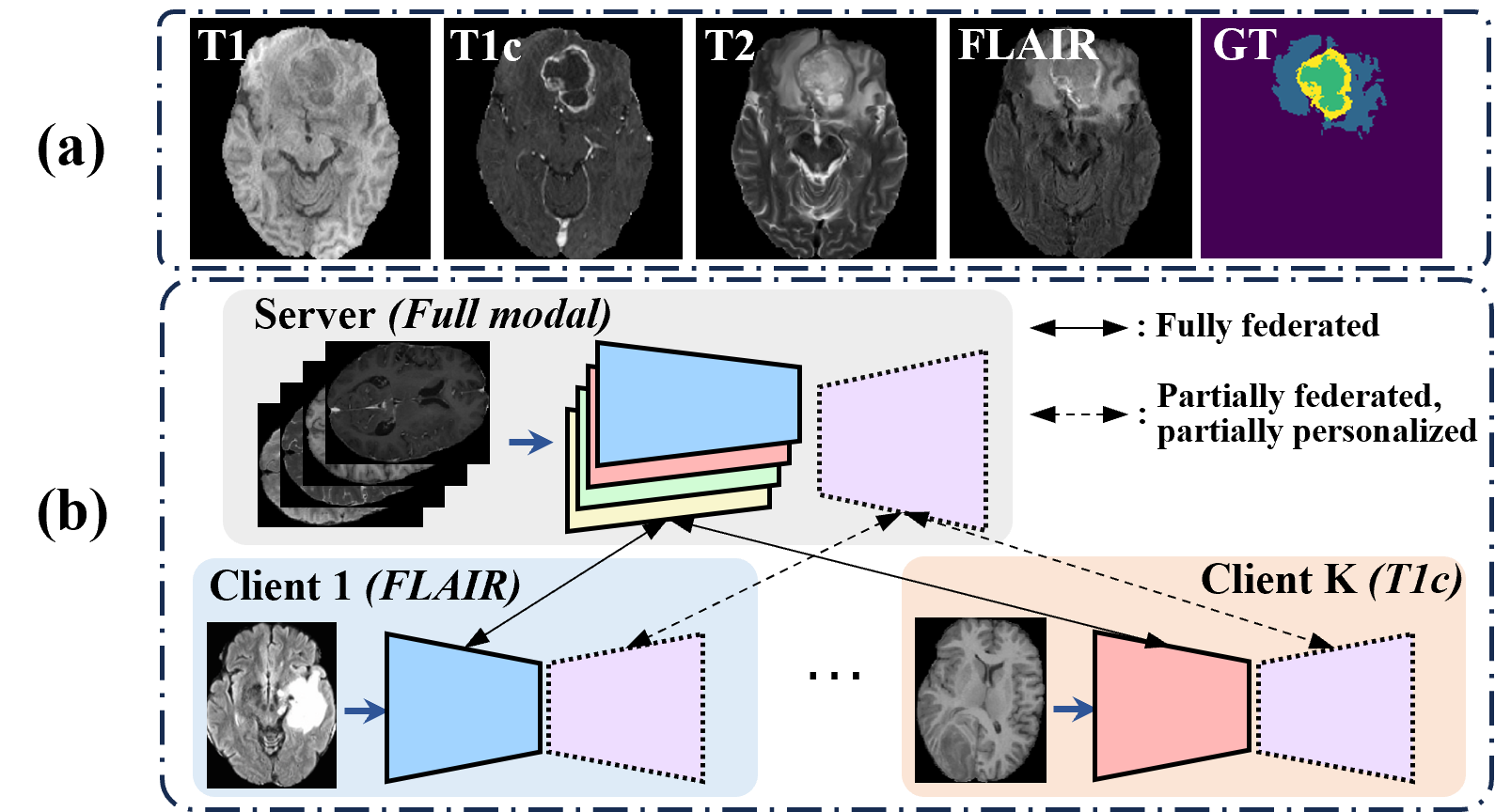} 
\caption{(a) Example images of the four modalities in BraTS 2020 \citep{menze2014multimodal} demonstrating severe intermodal heterogeneities, and corresponding tumor regions: blue: edema; yellow: enhancing tumor; and green: necrotic and non-enhancing tumor core.
(b) 
{\color{blue}To address the heterogeneities caused by different modal combinations across various sites, this work proposes adopting different aggregation strategies for modality-specific encoders (fully federated) and multi-modal fusion decoders (partially federated, partially personalized), to facilitate both common knowledge sharing and effective personalization.}
}
\label{fig:3-mri}
\end{figure}

One such application is brain tumor segmentation in multi-parametric magnetic resonance imaging (MRI) \citep{iv2018current}.
Specifically, four MRI modalities (in this work, we refer to MRI sequences as modalities following literature \citep{dorent2019hetero,menze2014multimodal,ding2021rfnet}) 
are commonly used to provide complementary information and support sub-region analysis: T1-weighted (T1), contrast-enhanced T1-weighted (T1c), T2-weighted (T2), and T2 fluid attenuation inversion recovery (FLAIR).
The first two modalities highlight the tumor core, and the last two highlight peritumoral edema (Fig. \ref{fig:3-mri}(a)).
When applying FL to such multimodal applications in practice, it is not uncommon that some participant institutes only possess a subset of the full modalities due to different protocols practiced, presenting a new challenge with the \textit{intermodal heterogeneity} across the FL participants.
In such a scenario, there can be two objectives for FL:
1) collectively training an optimal global model for full-modal input, and 
2) obtaining a personalized model for each participant \citep{chen2022fedmsplit,wang2019federated}, adapted for its data characteristics, and more importantly, better than trained locally without FL.
To our knowledge, these two objectives were rarely considered together in FL for medical image analysis. 

{\color{blue}This paper proposes a new FL framework for brain tumor segmentation with \textbf{fed}erated \textbf{m}odality-specific \textbf{e}ncoders and \textbf{p}artially \textbf{p}ersonalized multimodal fusion \textbf{d}ecoders (FedMEPD; Fig. \ref{fig:3-mri}(b)).}
Above all, to handle distinctively heterogeneous MRI modalities, FedMEPD employs an exclusive encoder for each modality, allowing a great extent of parameter specialization.
{\color{blue}In the meantime, while the encoders are fully shared between the server and clients, the decoders are partially shared and partially personalized dynamically, simultaneously catering to individual participants and common knowledge sharing in FL.}
Specifically, a multimodal fusion decoder on the server (\textit{i.e.}, a participant with full-modal data) fuses representations from the encoders to bridge the distribution gaps between modalities and reversely optimizes the encoders via backpropagation.
{\color{blue}Each client's fusion decoder is partially federated (and partially personalized) at the filter level based on the consistency between the parameter updates of the global and local models.
Intuitively, only those parameters whose updates consistently agree between the server and the client are federated.}
Meanwhile, multiple anchors are extracted from the fused multimodal representations at the server and distributed to the clients along with the shared parameters.
On the other end, the clients with incomplete modalities (including the special case of full-modal) calibrate their local missing-modal representations toward the global full-modal anchors via the scaled dot-product attention mechanism \citep{vaswani2017attention} to make up the information loss due to absent modalities and adapt representations of present ones.
To this end, we simultaneously obtain an optimal server model (for full-modal input) and personalized client models (for specific missing-modal input) from FL without sharing privacy-sensitive information.

In summary, our contributions are as follows:
\begin{itemize}
    \item We bring forward the intermodal heterogeneity problem due to missing modalities in FL for medical image analysis, and aim to obtain an optimal full-modal server model and personalized missing-modal client models simultaneously with a novel framework coined FedMEPD.
    %
    \item To tackle the intermodal heterogeneity, we propose to employ a federated encoder exclusive for each modality followed by a multimodal fusion decoder.
    \item {\color{blue}To simultaneously promote common knowledge sharing in FL and facilitate effective personalization, we propose to partially federate the fusion decoders based on the consistency between global and local parameter updates.}
    %
    \item In addition, we propose to extract and distribute multimodal representations from the server to the clients for local calibration of modality-specific features.
    \item Last but not least, we further enhance the calibration with multi-anchor representations.
\end{itemize}
Experimental results on the public BraTS 2018 and 2020 benchmarks \citep{menze2014multimodal,bakas2018identifying} show that our method achieves superior performance for both the server and client models to existing FL methods and that its novel designs are effective.

{\color{blue}This work substantially expands our preliminary exploration \citep{dai2024federated} in three main aspects.
First, we change the completely personalized multimodal fusion decoder to partially personalized and partially federated.
Such a change aims to promote the sharing of common knowledge in FL while still facilitating personalization, leading to notable improvements in clients' performance (\textit{cf.} Table \ref{tabs:ablation study}).
Second, in \citep{dai2024federated}, each client had data of only one modality, whereas clients in this work may have one to four (full) modalities for a more common and practical problem setting.
Third, we employ one more public benchmark dataset 
to demonstrate the effectiveness and generalization of the improved framework.}


%% file: sections/related_work.tex
\section{Related Work}

\subsection{Brain Tumor Segmentation with Multimodal MRI}

Multimodal MRI is the current standard of care for clinical imaging of brain tumors \citep{iv2018current}.
Segmentation and associated volume quantification of heterogeneous histological sub-regions are valuable to the diagnosis/prognosis, therapy planning, and follow-up of brain tumors \citep{menze2014multimodal}.
In recent years, deep neural networks (DNNs) significantly advanced state-of-the-art of brain tumor segmentation with multimodal MRI \citep{chen2020brain,chen2019dual,zhou2020one}. 
However, these methods were optimized for ideal scenarios where the complete set of modalities was present. 
In practice, scenarios in which one or more modalities are missing commonly occur due to image corruption, artifacts, acquisition protocols, contrast agent allergies, or cost constraints.
Therefore, many efforts have been made to accommodate the practical scenarios of missing modalities \citep{hu2020knowledge,wang2021acn,ding2021rfnet}. 
These methods successfully improved DNNs' feature representation capability against missing modalities---however, only in the centralized setting, limiting their efficacy in privacy-sensitive settings.
This work aims to address the missing-modal problem in the FL setting and eliminate the privacy issue.

\subsection{FL with Data Heterogeneity}

Data heterogeneity is a primary challenge in FL \citep{mcmahan2017communication}.
Personalized federated learning \citep[PFL;][]{tan2022towards} proposed to adapt the global model locally on clients' data to address this issue. 
{\color{blue}Existing works attempted improving the effectiveness of PFL by partial parameter sharing \citep{wang2022personalizing,li2021fedbn,collins2021exploiting,sun2021partialfed,wang2023feddp}, 
meta-learning \citep{t2020personalized,fallah2020personalized}, knowledge distillation \citep{yao2021local}, weight decay \citep{li2021ditto,zhang2023fedala,jiang2023iop}, and adaptive aggregation \citep{huang2021personalized,luo2022adapt,zhang2023grace}. 
These methods achieved decent performance in the presence of intramodal data heterogeneity, yet did not consider the intermodal heterogeneity of multimodal data.}
We are aware of several works for multimodal FL in the natural image domain.
\cite{xiong2022unified} proposed a co-attention mechanism to fuse the complementary information of different modalities, yet requiring all clients to have access to the same set of modalities.
FedIoT employed cross-modal autoencoders to learn multimodal representations in an unsupervised manner \citep{zhao2022multimodal}.
However, both methods \citep{xiong2022unified,zhao2022multimodal} only obtained a single global classifier without catering to the personalized needs of modal-heterogeneous clients.
\cite{yu2023multimodal} proposed a cross-modal contrastive representation ensemble between the server and hetero-modal clients by sharing a multimodal dataset, which may be unacceptable in strict privacy restrictions like healthcare.
In contrast, our framework optimizes a global model for full-modal input and simultaneously customizes a personalized model for each client's hetero-modal input.
It also maintains FL's data privacy by transmitting population-wise abstracted prototypes instead of image-wise features.


\subsection{Multimodal FL in Medical Image Analysis}

For medical images, heterogeneity issues in multimodal FL have yet to be thoroughly discussed.
FedNorm adapted the normalization parameters for different modalities while sharing common backbone parameters for computed tomography (CT)- and MRI-based liver segmentation \citep{bernecker2022fednorm}.
Yet, our experiments suggest that merely specializing in normalization parameters is insufficient to deal with the intermodal heterogeneity in multimodal brain tumor segmentation.
%
In this work, we propose handling the heterogeneity with modality-specific encoders to allow a greater extent of parameter specialization, followed by a multimodal fusion decoder to aggregate and fuse representations from the encoders and bridge the intermodal distribution gaps.

%% file: sections/2-method.tex

\section{Method}

\subsection{Problem Definition}

Denote the complete set of modalities by $M=$\{T1, T1c, T2, FLAIR\}, which is indexed by $m$, and a full-modal input by $X_M\in \mathbb{R}^{|M|\times D \times H\times W}$, where $D$, $H$, and $W$ are the depth, height, and width of the volume, respectively.
We consider a heterogeneous FL setting as described below.
A server has access to a full-modal dataset $\mathcal{D}_M = \{(X_M, Y)\}$, where $Y \in \mathbb{R}^{N_c \times D \times H \times W}$ is the segmentation mask and $N_c$ is the number of target classes.
The server coordinates several clients with access to datasets of potentially incomplete modalities (``missing-modal'') denoted by $\mathcal{D}_{\{m\}}^i=\{(X_{\{m\}}, Y)\}$, where $i$ indicates the $i$\textsuperscript{th} client and $X_{\{m\}} \in \mathbb{R}^{|\{m\}| \times D \times H\times W}$.
We will omit the index $i$ when there is no risk of confusion.
{\color{red}The server may be a major regional hospital, whereas the clients may be smaller local health units.}
Note that we do not constrain the clients' modalities for practical consideration. 
Thus, $\mathcal{D}_{\{m\}}$ can have one to four modalities, \textit{i.e.}, $1 \leq |\{m\}| \leq |M|$, corresponding to single- to full-modal datasets.



This work aims to train, via FL, not only an optimal global model that works well with full-modal data but also optimal personalized models that work well in specific missing-modal situations for the clients.
The latter is practically meaningful as it is usually difficult for a local health unit to upgrade its imaging protocol quickly, yet still look forward to training an optimal model for its current protocol by participating in FL. 
\input{sections/figs/fig_overview.tex}
\subsection{Framework Overview}
Without loss of generality, we use four clients, each with data of a mutually different modality, for method description.
As shown in Fig.~\ref{fig:2-overview}, the server has four modality-specific encoders (one for each modality) and a modal fusion decoder, whose fused features are clustered to produce multimodal anchors.
{\color{blue}Meanwhile, each client has a fully federated modality-specific encoder for each modality and a partially federated, partially personalized fusion decoder.}
%
Additionally, a localized adaptive calibration via cross-attention (LACCA) module calibrates the clients' missing-modal representations toward the server's multimodal anchors.

\subsection{Federated Modality-specific Encoders}
In classical FedAvg \citep{mcmahan2017communication}, the server and clients usually share the same network architecture, where the server aggregates and averages the network parameters of the clients and then distributes the averaged parameters back to the clients in a straightforward manner.
However, due to the high heterogeneity among the multimodal MRI data (see Fig.~\ref{fig:3-mri}),
our problem setting becomes challenging for this paradigm.
Instead, we propose federated modality-specific encoders to handle the distinctively heterogeneous imaging modalities.
On the one hand, we adopt an 
architecture with late fusion strategy \citep{ding2021rfnet} to compose the global model on the server, including a modality-specific encoder $E_m$ (parameterized by $W_m^s$) for each modality, a fusion decoder $D_M$ (parameterized by $W_d^s$) for multimodal feature aggregation and fusion, and a regularizer (not shown in Fig. \ref{fig:2-overview} for simplicity).
The regularizer is a straightforward auxiliary segmentation decoder shared by all modality-specific encoders. 
It regularizes the encoders to learn the same discriminative features by forcing them to share the decoder parameters.
Please refer to \citep{ding2021rfnet} for details.\footnote{Note that our framework is model-agnostic and can be implemented with various non-FL multimodal segmentation models consisting of modality-specific encoders and multimodal fusion decoders, \textit{e.g.}, \citep{dorent2019hetero,shen2019brain,zhou2021latent}.
In this work, we use \citep{ding2021rfnet} for demonstration purposes due to its outstanding performance and straightforward architecture.}
Given a full-modal input $X_M$, each $E_m$ first extracts features from the corresponding modality $X_m$, followed by $D_M$ fusing multimodal features and generating segmentation masks.
On the other hand, each client has a federated modality-specific encoder $E_m$, and a partially federated, partially personalized decoder $D_m$ for generating segmentation masks.
$E_m$ on the clients shares the same architecture as the server.

In each round of FL, the clients first receive parameters $W_m^s$ from the server to replace its local copy $W_m^i$, where $i\in\{1,\ldots,N_m\}$ and $N_m$ is the number of clients with data of modality $m$, train for $N_e$ epochs on local data, and then send updated $W_m^i$ back to the server.
After receiving $W_m^i$, the server averages $W_m^i$ of the same modality: 
\begin{equation}\label{encoder_agg}\small
    W_m^s = \frac{1}{N_m}{\sum}_i W_m^i,
\end{equation}
trains for $N_e$ epochs on its full-modal data, and sends updated $W_m^s$ to the clients for the next round.
To this end, the server effectively bridges the distribution gaps between the modalities with the fusion decoder $D_M$. 
It utilizes complementary multimodal information to train each modality-specific encoder $E_m$ via backpropagation.

\subsection{Partially Personalized Fusion Decoder}\label{sec:partial}

{\color{blue}Several works proposed to split the model into feature extractor and classifier for classification tasks \citep{collins2021exploiting,wang2022personalizing}, with the latter typically personalized in PFL.
Similarly, our previous work split the segmentation model into modality-specific encoders and a multimodal fusion decoder, with the latter personalized and trained only using local data \citep{dai2024federated}.
However, unlike the classification head, the decoder's parameters account for a significant portion of the segmentation model.
Entirely personalizing the decoder could impede general knowledge sharing among participants and undermine the benefits of FL.
In addition, it may also lead to overfitting for participants with limited local data.
Meanwhile, due to the heterogeneous modal distributions of clients' data, directly aggregating the decoder parameters would also degrade performance.

To address the above issues, we propose a novel PFL method that leverages the consistency between the global and client parameter updates to guide a partial personalization of the decoder.
After $r-1$ rounds of FL, let us denote the decoder parameters at the $i$\textsuperscript{th} client by $W^{i,r-1}_d$, and the global decoder parameters of the server by $W_d^{s,r-1}$. 
Then, at the beginning of the $r$\textsuperscript{th} round, the partially personalized, partially federated parameter aggregation rule is formulated as
\begin{equation}\label{equ:local_update}\small
    W_d^{i,agg} = (1-B^{i,r-1})W_d^{i,r-1} + B^{i,r-1}W_d^{s,r-1},
\end{equation}
where $B^{i,r-1}$ is a binary indication mask: 1 for federating (using global parameters) and 0 for personalizing (using local parameters).
We will discuss the dynamic update of $B^i$ shortly.
After the aggregation, $W_d^{i,agg}$ is trained locally on the client's data for $N_e$ epochs, yielding $W_d^{i,r}$.

For the decoder on the server, the partially federated parameter aggregation rule is formulated with an exponential moving average~\citep[EMA;][]{tarvainen2017mean} strategy to balance the contributions of the server and clients:
\begin{equation}\label{decoder_agg}\small
    W_d^{s,agg} =\lambda  W_d^{s,r-1} + (1-\lambda)  W_d^{loc,r},
\end{equation}
where $\lambda$ is a hyper-parameter, and
\begin{equation}\label{aggregation}
 W_d^{loc,r} = \frac{1}{\sum_{i}B^{i,r-1}}{\sum}_{i} B^{i,r-1} H^{i,r} W_d^{i,r},
\end{equation}
where $H^{i,r}$ is a normalization term that is inversely correlated with the magnitude of the difference between $W_d^{i,r}$ and $W_d^{i,agg}$.
It is designed to mitigate the client bias \citep{wang2020tackling}: some clients may have larger updates to $W_d^{i,agg}$, leading to disproportionately greater influence in Eqn. (\ref{aggregation}).
Intuitively, the server-end aggregation is an EMA of the server's parameters and a selective aggregation of the clients'.
Similar to the clients, the server trains $W_d^{s,agg}$ for $N_e$ epochs on its own data, yielding $W_d^{s,r}$. 

To update $B^{i}$ dynamically, we compute the round-wise updates to the decoder parameters of the server and clients:
\begin{equation}  \small
    \Delta W_d^{s,r} = W_d^{s,r} -  W_d^{s,r-1}\ \text{and}\ \Delta W_d^{i,r} = W_d^{i,r} -  W_d^{i,agg},
\end{equation}
respectively.
Then, we use the consistency between the server's and clients' updates to guide the personalization of the decoder parameters. 
Concretely, the \textit{per-filter} cosine similarity between $\Delta W_d^{s,r}$ and $\Delta W_d^{i,r}$ is calculated: $\delta^{i,r}_j = \operatorname{cos}(\Delta \mathbf{w}_j^{s,r}, \Delta \mathbf{w}_j^{i,r})$, where $\Delta\mathbf{w}_j^{s,r}$ and $\Delta\mathbf{w}_j^{i,r}$ are the parameter updates to the $j$\textsuperscript{th} filter at the server and the $i$\textsuperscript{th} client, respectively.
In concept, the $j$\textsuperscript{th} filter is personalized 
if $\delta^{i,r}_j < 0$ for $P$ consecutive rounds, or federated otherwise.
Formally, we have

\begin{equation}\label{eq:fed_or_per}
    b_j^{i,r} = \begin{cases}
        0, & \text{if } b_j^{i,r-1} = 0, \\
        0, & \text{if } \delta^{i,r}_j < 0\ \text{for } P \text{ consecutive rounds}, \\
        1, & \text{otherwise},
    \end{cases}
\end{equation}
where $b_j^{i,r} \in B^{i,r}$ is a binary scalar indicating the federating/personalizing status of the $j$\textsuperscript{th} filter, and $P$ is a hyper-parameter referred to as \textit{patience}.
$P$ regulates the extent of personalization: the smaller $P$ is, the more personalized are the clients' decoders.
Note the first case in Eqn. (\ref{eq:fed_or_per}) states that once a filter is personalized in a certain round, it will stay personalized for the rest of the FL.
This is to avoid jumping back and forth in federating and personalizing status, which may lead to unstable training or even failure.

Intuitively, suppose the update direction of a client filter aligns with its counterpart on the server.
In that case, it positively contributes to the global model update and should, therefore, be included in the global model aggregation.
Conversely, suppose it consistently updates in a direction that contradicts the global one.
In that case, this filter is potentially sensitive to the data discrepancy between the client and the server, suggesting that it should be personalized for further training.
We employ filters as the basic unit of personalization for two reasons.
First, filters in convolutional networks often learn to detect specific features \citep{zeiler2014visualizing}. 
By personalizing at the filter level, we maintain the integrity of these learned features, leading to consistent and stable personalization.
Second, the mask $B^i$ needs to be transmitted from the server to the client.
By using filters as units, we can 
limit the communication overhead to a low level, 
since only a single byte is needed to indicate a filter's personalizing/federating status.

}

For implementation, we initialize $B^{i,0}$ to all ones.
For Eqn. (\ref{aggregation}), $\eta_j^{i,r} \in H^{i,r}$ is inversely correlated with $\big|| \Delta \mathbf{w}_{j}^{i,r} \big||_2$, subject to $\sum_i \eta_j^{i,r} = 1$.
{\color{red}For Eqn. (\ref{decoder_agg}), $\lambda$ is dynamically determined according to the personalizing status of a filter:
\begin{equation}
    \lambda = \begin{cases}
        1, & \text{if } b_j^{i,r-1} = 0 \text{ for all } i, \\
        0.3, & \text{otherwise}.
    \end{cases}
\end{equation}
In other words, $\lambda$ is set to 1 if the $j$\textsuperscript{th} filter is completely personalized for all clients (and thus the server) so that $W_d^{s,agg}$ is entirely updated by $W_d^{s,r-1}$ and no scaling occurs for $W_d^{s,r-1}$, or empirically set to 0.3 otherwise.
}

\subsection{Multi-Anchor Multimodal Representation} 
\label{2-2}
Besides aligning modality-specific encoders with the multimodal fusion decoder, the server also learns multi-anchor multimodal representations for the classes of interest.
The representations are distributed to the clients in addition to the encoder and decoder parameters.
\cite{liu2020federated} proposed communicating encoded input features directly, which may breach the privacy restriction of FL.
On the contrary, some works proposed to transmit category prototypes \citep{mu2023fedproc,tan2022fedproto}.
Yet a single prototype was highly compressed and may not carry enough representative information for a class, especially considering the significant inter-subject variations in 3D multimodal medical images.

In this work, we propose to extract \textit{multiple} prototypes \citep{cui2020unified} from the \textit{fused multimodal} features for each class of interest for enhanced representation power, which we refer to as anchors for their calibration purpose \citep{ning2021multi}.
Concretely, we extract per-class features from the fused multimodal feature maps of the decoder $D_M$ by masked average pooling using the ground truth segmentation, and apply the K-means method \citep{macqueen1967classification} to the extracted features to obtain $N_k$ anchors.
For the $l$\textsuperscript{th} feature scale level, where $l\in\{1,\ldots,4\}$ for the networks we use, the anchors for all the $N_c$ classes can be collectively denoted by $A_l\in\mathbb{R}^{N_k N_c \times C_l}$, where $C_l$ is the number of feature channels.
Empirically, we determine the cluster membership using the most abstract feature level, \textit{i.e.}, $l=4$, and apply the membership to compute $N_k=4$ anchors for all levels (see Table \ref{tabs:clusNum} for corresponding experiments).
This strategy can preserve the full-modal information of each class well while incurring little network transmission burden. 
{\color{red}Meanwhile, the few class-wise anchors are abstracted from the entire training population on the server, thus carrying little private information concerning individuals.}%
\footnote{\color{red}We caution that besides individual privacy of the patients' image data, other information might also face potential privacy risks during the FL.
Our framework is compatible with typical privacy protection algorithms for FL. 
However, as this work is focused on addressing the intermodal heterogeneity in multimodal medical imaging FL, we refer readers to the thorough review of privacy preservation in FL for healthcare by \cite{pati2024privacy}.}
To avoid the collapse of the training process due to jumps in cluster centroids as a result of re-clustering at each round \citep{xie2016unsupervised}, we treat the anchors as a memory bank and update them smoothly via EMA~\citep{tarvainen2017mean}: $\Bar{a}_c = \omega\Bar{a}_c + (1-\omega)a_c$, 
where $\Bar{a}_c$ is an anchor for class $c$ in the memory bank and updated by the closest cluster centroid $a_c$ of the same class, and $\omega$ is a smoothing coefficient.
We set $\omega$ to 0.999 following \cite{tarvainen2017mean}.


\begin{figure}[t]
\centering
\includegraphics[trim=0 5 0 5.5, clip,width=.7\textwidth]{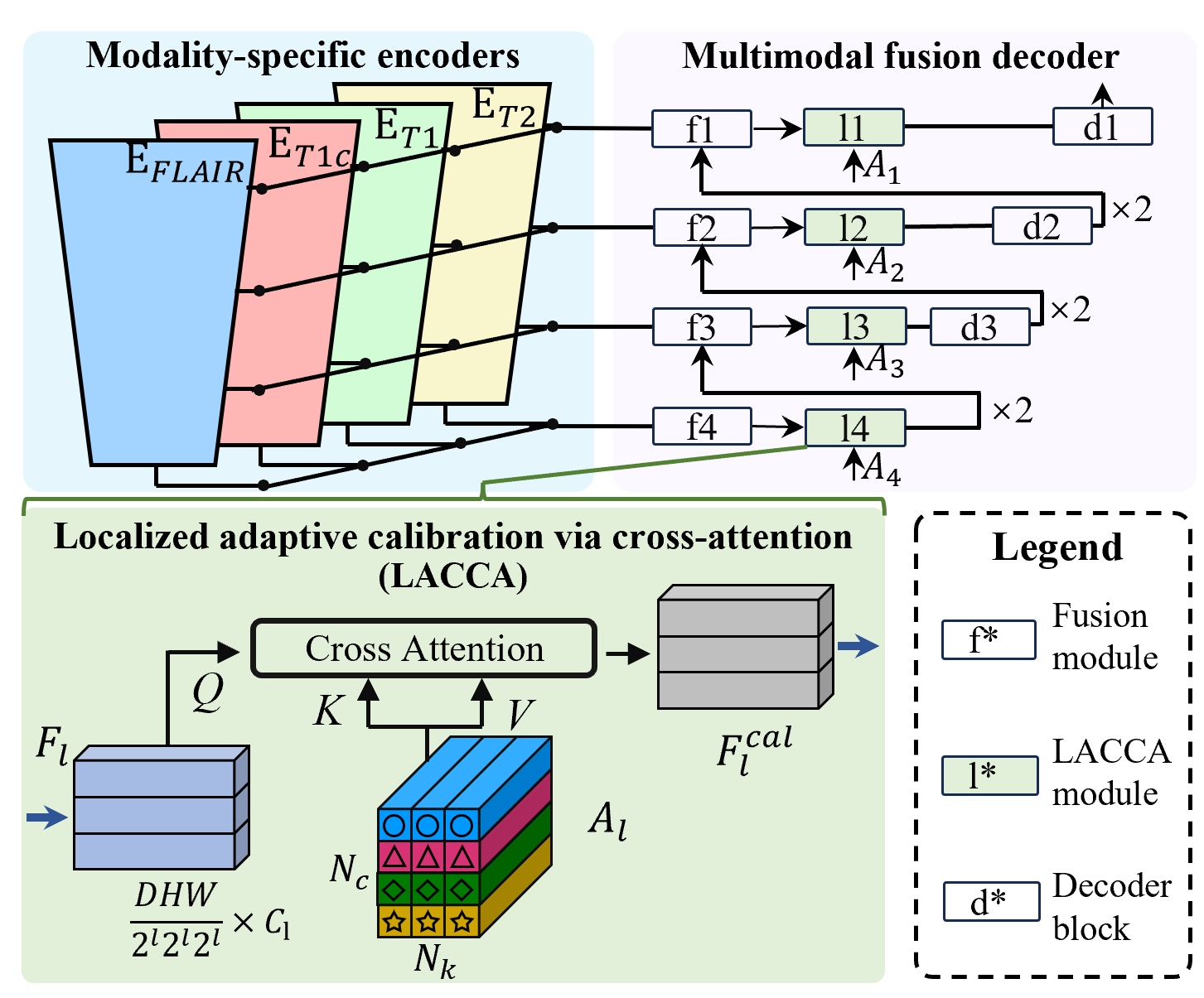} 
\caption{\color{blue}Illustration of the proposed localized adaptive calibration via cross-attention (LACCA) module.
}
\label{fig:lacca}
\end{figure}

\subsection{Localized Adaptive Calibration via Cross-Attention}\label{2-3}

In our framework, the multimodal fusion decoders at the server and various clients must fuse features of vastly different modal combinations.
{\color{blue}To handle such significant intermodal heterogeneity, we have proposed partially personalizing the decoder parameters in Section \ref{sec:partial}.}
In this section, we further propose to calibrate the clients' local missing-modal representations toward the global full-modal anchors to make up for the information gap caused by absent modalities.


In each federated round, the clients receive the multimodal anchors $A_l$ from the server and use the anchors to calibrate local missing-modal representations.
Concretely, as shown in Fig. \ref{fig:lacca}, denoting the input feature map to the $l$\textsuperscript{th} scale level of the decoder by $F_l\in\mathbb{R}^{\frac{D}{2^l} \times \frac{H}{2^l} \times \frac{W}{2^l} \times C_l}$,
we reshape $F_l$ to the dimension $\frac{DHW}{2^l2^l2^l} \times C_l$.
Then, inspired by the attention operation in the Transformer architecture \citep{vaswani2017attention}, we treat the reshaped $F_l$ as queries and the multimodal anchors as the keys and values, and calibrate the local representations toward the global multimodal anchors by the cross attention: 
\begin{equation}\label{eq}\small
    F_l^\mathrm{cal} 
    = \operatorname{Attn}(F_l, A_l)
    = \operatorname{softmax}\big[{F_lW_0 (A_lW_1)^T}\big/{\sqrt{C_l}}\big]A_lW_2,
\end{equation}
where $W_0$, $W_1$, and $W_2$ are linear projection matrices.
Finally, the calibrated features $F_l^\mathrm{cal}$ are reshaped back to participate in subsequent forward propagation.
The calibration process is localized and self-adaptive in that each client locally emphasizes the part of the global multi-anchor multimodal representations that best suits its own data modality and distributions---via the dot-product attention---to yield more powerful models tailored for itself.
To this end, we name it the \textit{localized adaptive calibration via cross-attention} (LACCA) module.
The LACCA module is inserted in all four feature scales of our backbone networks.
Note that the multimodal anchors are learnable parameters during FL, and are directly used by the clients for inference after training.
The complete algorithm of our method is provided in Algorithm \ref{algori1}.

\begin{algorithm}[!t] 
    \caption{FedMEPD algorithm.} \label{algori1}
    \begin{algorithmic}[1]\small
    \Require the modality set $M$ = \{T1, T1c, T2, FLAIR\} indexed by $m$, a full-modal training dataset $\mathcal{D}_M$ on the server, the number of clients $N$, the missing-modal training set $\mathcal{D}^i_{m}$ on client $i$,
    the number of communication rounds $N_r$, the number of training epochs $N_e$ in each round, and the set of federating/personalizing masks $\{B^{i}\}$ for the decoders.
    
    \renewcommand{\algorithmicensure}{\textbf{Output}:} 
    \Ensure the collection of encoder parameters $W^s_{\{m\}}=\{W^s_m\}$, the server decoder parameters $W_d^s$, the set of client decoder parameters $\{W^{i}_d\}$, and the collection of multimodal anchors $A_{\{l\}}=\{A_l\}$ for different feature scale levels $l$.
    \vspace{1ex}
        \renewcommand{\algorithmicfunction}{\textbf{Server executes}:} 
        \Function{}{}
        \State \parbox[t]{\dimexpr\linewidth-\algorithmicindent}{Initialize $W_{\{m\}}^s$, $W_d^s$, $\{B^i\}$, and update $W_{\{m\}}^s$, $W_d^s$ by training on $\mathcal{D}_M$ for $N_e$ epochs}
        \State Initialize $A_{\{l\}}$ by K-means
        \For {round $r=1$ to $N_r$}
            \For {each client $i\in N$}
            \State $W_{m}^{i}, W_d^{i} \gets$ ClientUpdate$(m, i, W_{m}^{s}, W_{d}^s, A_{\{l\}}, B^i)$ 
            \EndFor
            \For {$m \in M$}
            \State \parbox[t]{\dimexpr.8\linewidth-\algorithmicindent}{Aggregate parameters $W_m^s$ for modality-specific encoder according to Eqn. (\ref{encoder_agg})}
            \EndFor
            \State \parbox[t]{\dimexpr.94\linewidth-\algorithmicindent}{Aggregate decoder parameters $W_{d}^s$ according to Eqn. (\ref{decoder_agg}) and Eqn. (\ref{aggregation})}
            \State Update $W_{\{m\}}^s$, $W_d^s$ by training on $\mathcal{D}_M$ for $N_e$ epochs
            \State Update $A_{\{l\}}$ by exponential moving average
            \State Update $\{B^{i}\}$
        \EndFor\EndFunction\vspace{1ex}

        \renewcommand{\algorithmicfunction}{\textbf{ClientUpdate}($m,i,W^s_m,W_{d}^s,A_{\{l\}},B^i$):} 
        \Function{}{}\Comment{run on client $i$ with modality $m$}
        \State $W_m^i \gets W_m^s$ 
        \State Aggregate $W_d^i$ with $W_{d}^s$ and $B^i$  according to Eqn. (\ref{equ:local_update})
        \State \parbox[t]{\dimexpr\linewidth-\algorithmicindent}{Update $W_m^i$ and $W_d^i$ by training on $\mathcal{D}_m^i$ with LACCA (Eqn. (\ref{eq})) for $N_e$ epochs}
        \State return $W_m^i$, $W_d^i$
        \EndFunction
    \end{algorithmic}
\end{algorithm}

%% file: sections/figs/fig_overview.tex
\begin{figure*}[t]
\centering
\includegraphics[trim=0 2 0 3, clip, width=.98\textwidth]{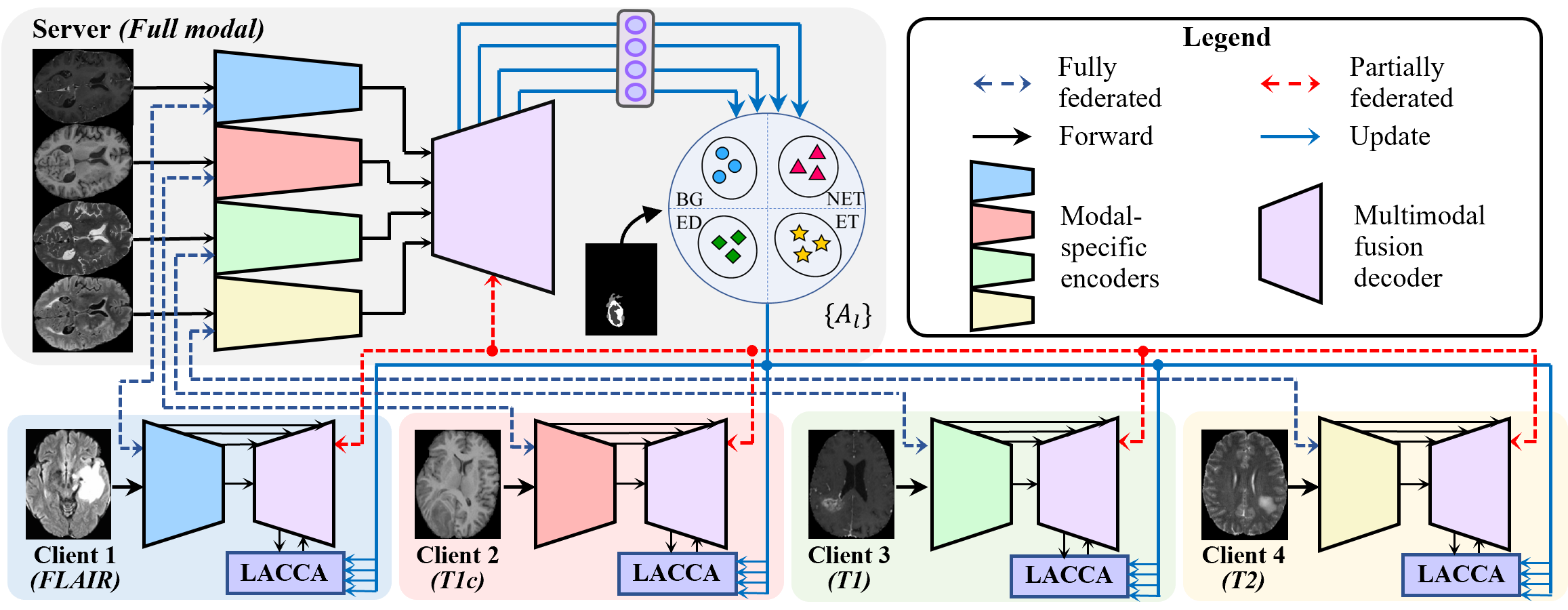} 
\caption{{\color{red}Overview of the proposed FedMEPD framework.}
{The server has four modality-specific encoders (one for each modality) and a multimodal fusion decoder, whose fused features are clustered to produce multimodal anchors.
Each client has a fully federated modality-specific encoder for each modality and a partially federated (partially personalized) fusion decoder.
A localized adaptive calibration via cross-attention (LACCA) module calibrates the clients’ missing-modal representations toward the server’s multimodal anchors.}
}
\label{fig:2-overview}
\end{figure*}

%% file: sections/3-experiments.tex
\section{Experiments and Results}

\subsection{Datasets and Experimental Setting}
We conduct experiments on the multimodal Brain Tumor Segmentation (BraTS) 2018 and 2020 datasets~\citep{menze2014multimodal,bakas2018identifying}. 
The two datasets consist of 285 and 369 multi-contrast MRI scans, respectively, with four sequences: T1, T1c, T2, and FLAIR.
The goal is to segment three nested subregions of brain tumors: whole tumor, tumor core, and enhancing tumor.
In our problem setup, we first partition the entire training set into nine distinct, non-overlapping subsets. 
This is done by ensuring that all cases sourced from the same institute are in the same subset, per BraTS's official information.
One of the nine subsets is designated as the server-end dataset, while the remaining eight are client datasets.
We assign different modal combinations, from mono- to full-modal, to the clients.
This results in two clients for each of mono-, dual-, triple-, and full-modal combinations.
{\color{blue}
For BraTS 2018, the numbers of subjects assigned to the clients are: mono-modal---21 and 22, dual-modal---20 and 22, triple-modal---21 and 22, and full-modal---34 and 34, and those to the server are 88.
For BraTS 2020, the numbers of subjects assigned to the clients are: mono-modal---30 and 26, dual-modal---28 and 22, triple-modal---34 and 31, and full-modal---34 and 35, and those to the server are 129.}
Each of these subsets is further divided into training, validation, and testing sets according to the ratio of 6:2:2.\footnote{We will publish the data partition with our code.}
Each client can only access the specific modalities assigned to it.

We use the mean Dice similarity coefficients (mDSC) {\color{red}and the 95\textsuperscript{th} percentile of the Hausdorff distance (HD95)} of the three tumor subregions as the evaluation metrics, and the Wilcoxon signed-rank test to analyze statistical significance.
For experiments, we use FLAIR (F), T1c (C), T1, and T2 to indicate clients with corresponding modalities, Avg to indicate the clients' average performance, and S to indicate the server.


\subsection{Implementation}
The proposed framework is implemented using PyTorch (1.13.0) and trained with five RTX 2080Ti GPUs, with the server on one GPU and the clients evenly distributed on the rest.
We use the RFNet \citep{ding2021rfnet} as our server network.
The client networks are the same as the server, except they only have the modality-specific encoders corresponding to the clients' local data modalities.
The LACCA module is implemented with eight attention heads.
The input crop size is $80\times80\times80$ voxels, and the batch size is set to 1 for the server and clients.
The commonly used Dice loss \citep{milletari2016v} plus the cross entropy loss for medical image segmentation are employed.
The Adam optimizer, with its learning rate and weight decay set to 0.0002 and $10^{-5}$, respectively, is leveraged for optimization.
We train the networks for 1000 rounds, and in each federated round, the server and clients are trained for one epoch.
We follow \cite{ding2021rfnet} for data preprocessing and augmentation.
{\color{red}We have released our code, data split, and trained models at https://github.com/ccarliu/FedMEPD.}

\subsection{Comparison with Baseline and State-of-the-Art (SOTA) Methods}
We compare our proposed FedMEPD to various baseline and SOTA FL algorithms.
As the baseline, the server and client models are trained locally on the private data of each site.
As to FL algorithms, we adopt the classical FedAvg \citep{mcmahan2017communication}, several up-to-date approaches to FL on multimodal data or personalized models, including FedMSplit \citep{chen2022fedmsplit}, personalized FL (perFL)~\citep{wang2019federated}, IOP-FL \citep{jiang2023iop}, FedIoT \citep{zhao2022multimodal}, CreamFL \citep{yu2023multimodal}, FedNorm \citep{bernecker2022fednorm}, {\color{red}and two top-ranked methods in Federated Tumor Segmentation Challenge 2021 and 2022 \citep{pati2021federated,baid2021rsna}: FedCostWAvg \citep{machler2021fedcostwavg} and FedPIDAvg \citep{machler2022fedpidavg}}.
Under the premise of keeping the methodological principles unchanged, necessary adaptations are made to ensure fair comparison in our practical experimental settings: 
1) for FedAvg and its derived methods (perFL, IOP-FL, FedNorm{\color{red},  FedCostWAvg, and FedPIDAvg}), 
which did not originally conduct server-end training, we make them do so on the server data like our method,\footnote{Our preliminary experiments empirically showed that with the server-end training, they performed better than without it.} and 
2) we change FedIoT's autoencoding clients to supervised networks.
Also, we use the same backbone networks as our mono-modal clients' for FedAvg and derivatives (in FedAvg architecture, the server and clients use the same network structure), and the same networks as our server and clients for the counterparts in other methods.
It should be noted that as CreamFL requires sharing the server data with all clients, 
it violates the privacy restriction in the medical context and increases the training data for the clients.
Lastly, as RFNet \citep{ding2021rfnet} was originally designed for both full- and missing-modal segmentation after training with full-modal data, we also train it on the server data in its original recipe and evaluate its performance for comparison.

\input{sections/tabs/tab_test1}

\input{sections/tabs/tab9_hd95}

\begin{figure*}[!t]
\centering
\includegraphics[trim=0 4 0 0, clip,width=\textwidth]{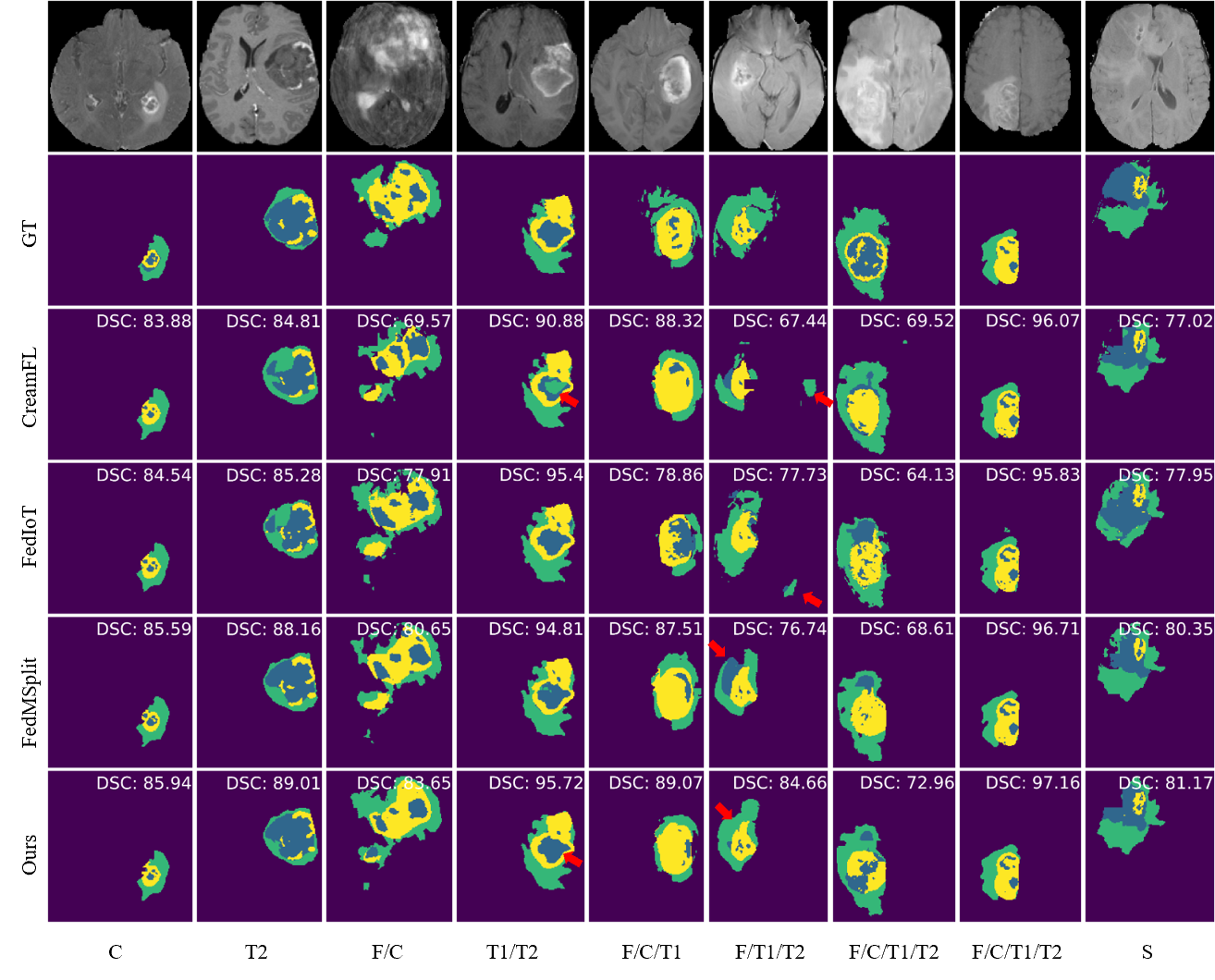} 
\caption{{\color{red}Example segmentation results of BraTS 2018 for a random subject in the test set of each site, by our method and the other three best-performing methods in terms of the average client mDSC in Table \ref{tabs:state-of-art}.
The red arrows highlight the advantages of our method.}
The Dice similarity coefficients (\%) are overlaid for reference.
Blue: necrotic and non-enhancing tumor core, yellow: enhancing tumor, and green: edema.
}
\label{fig:Setting1}
\end{figure*}

\input{sections/tabs/tab5.tex}

The mDSC results are shown in Table~\ref{tabs:state-of-art}.
As the comparative trends are similar for both datasets, we mainly describe the results on BraTS 2018 below.
{\color{blue}With none or limited modality-specific parameters, FedAvg and its derived methods for PFL (perFL, FedNorm, IOP-FL, {\color{red}FedCostWAvg, and FedPIDAvg}) mostly yield worse performance than the baseline local models.
This indicates that the severe intermodal heterogeneity impedes the classical FL architecture from effectively utilizing the extra data on the clients.
%
In contrast, the specialized multimodal FL frameworks, CreamFL, FedIoT, and FedMSplit, achieve comprehensive improvements;
especially, FedMSplit excels in the clients' average mDSC by $\sim$12\% compared with FedAvg.
%
Our proposed method further improves upon FedMSplit by $\sim$5\% in the clients' average mDSC, and achieves the best performance for all clients and the server.}
Remarkably, besides improving the clients' performance with locally personalized models, our method also substantially enhances the performance of the server by effectively exploiting the clients' data of heterogeneous modalities.
Lastly, the RFNet, despite being a strong model for various missing-modal situations, only yields performance comparable to the baseline local models, probably because it does not make use of the extra data on clients.
We visualize example segmentation results in Fig. \ref{fig:Setting1}.
The comparative trends are similar to Table \ref{tabs:state-of-art}.

{\color{red}Additionally, Table \ref{tabs:state-of-art_hd} charts the HD95 results.
As we can see, the comparative trends are similar to the mDSCs in Table \ref{tabs:state-of-art}: our method again achieves the best performance for the server-end HD95 and the average client HD95, as well as for most client-wise HD95. 
These results provide another perspective for a more comprehensive evaluation of our method's effectiveness in hetero-modal medical image FL.
}

\subsection{Validating Designs for Multi-Anchor Multimodal Representations}
Based on the validation data, we first determine 1) the optimal number of multimodal anchors per class ($N_k$) and 2) the optimal feature scale level $l$ based on which the cluster membership is determined.
We fix either to reduce the search space while varying the other.
The results in Table \ref{tabs:clusNum} look reasonably stable.
We select $N_k=4$ and $l=4$ for evaluation and comparison with other methods on the test data due to their highest performances in both the clients' average and the server's mDSCs compared with alternative values.
We conjecture that $l=4$ (\textit{i.e.}, the most abstract feature level) works the best due to its excellent capability of abstraction and denoising despite the relatively low resolution.

\input{sections/tabs/tab6.tex}

{\color{blue}
\subsection{Sensitivity to Personalization Patience}

The multimodal fusion decoder's extent of personalization is controlled with a hyper-parameter $P$ (Eqn. (\ref{eq:fed_or_per})).
We study FedMEPD's sensitivity to varying $P$ values and show the results in Table \ref{tabs:patience}.
On the one hand, complete personalization without federating any parameters ($P=0$) yields the worst results in both the clients' average and the server's mDSCs.
Therefore, overlooking the common knowledge sharing would lead to inferior performance even in the presence of severe intermodal heterogeneity.
On the other hand, excessive parameter federation with larger $P$'s also degrades the performance slightly.
Notwithstanding, the performance of the server and clients is fairly stable with $P=6$--$14$.
We select $P=10$ for test-set evaluation and comparison with other methods.
}

\subsection{\color{red}Impact of Server-End Full-Modal Data}

{\color{red}
As the multimodal anchors are derived from the full-modal data on the server, we conduct two more experiments with the BraTS 2018 dataset to investigate how the quantity and quality of the server’s data would affect the performance of our framework.

First, we have experimented with reducing the server data to 50\%, 30\%, and 10\% of the original amount. 
The results are shown in Table \ref{tabs:server_limit}.
As we can see, the average client mDSC only drops marginally to slightly, from 75.70 (all data) to 74.34 (50\% data), 73.81 (30\% data), and 72.81 (10\% data), which is desirable.
Even with only 10\% server data, the average client mDSC of our method is still better than all compared ones with all server data: 72.81 (ours with 10\% server data) versus 71.23 (FedMSplit, the second-best method in Table~\ref{tabs:state-of-art} in terms of average client mDSC, with all server data). 
In addition, it is worth noting that 30\% and 10\% server data (about 26 and 9 samples) is actually less than that of the two full-modal clients (34 samples each).
In practice, better overall performance may be achieved by switching the server role to one of the two full-modal clients.
Meanwhile, the server's performance drops more obviously due to the substantially reduced training data, which is reasonable.

Second, we have also experimented with reducing the quality of the server data by randomly dilating or eroding every annotation mask by one pixels.
The results are shown in Table \ref{tabs:server_limit}, too.
Favorably, the average client mDSC only drops marginally to 75.02, without statistical significance from that obtained with the original annotation masks.
Note that the server performance drops more noticeably due to the fact that the training masks on the server are directly corrupted and thus already diverge from the ground truth.


To summarize, these results indicate that our proposed multi-anchor multimodal representation and localized adaptive calibration are not only effective in multimodal representing and calibrating, but also efficient and robust with respect to the quantity and quality of the full-modal server data.
}

\input{sections/tabs/tab8_limit_server}

\input{sections/tabs/tab_valid}

\subsection{Ablation study}
We conduct thorough ablative experiments to validate the efficacy of our novel framework design, including the federated modality-specific encoders, the partially federated, partially personalized fusion decoders, the LACCA module, and the multi-anchor multimodal representations.
The results are presented in Table~\ref{tabs:ablation study}.
The first two rows are variants of the classical FedAvg \citep{mcmahan2017communication} with either the encoder or decoder federated, and others are variants of our proposed method.
Also, row (g) is our preliminary exploration \citep{dai2024federated}.
We can see that of the variants of FedAvg, federating the decoder while personalizing the encoder (row (b)) outperforms the reverse (row (a)).
In comparison, our federated modality-specific encoders (row (c)) achieve substantial performance boosts over the FedAvg family, \textit{e.g.}, the clients' average mDSC improves by $\sim$4\% compared with row (b).
{\color{blue}However, simply federating the multimodal fusion decoder (row (d)) slightly degrades the clients' average mDSC.
In contrast, row (e) adopts our proposed partially federating, partially personalizing strategy for the decoder, achieving notable improvements in both the clients' average and the server's performance.}
Row (f) additionally incorporates the LACCA module but with mono-anchor representations obtained by averaging all server data, achieving further improvements for the clients.
Row (h) is our full model with multi-anchor representations, yielding the highest performance in both the average missing-modal performance across the clients (72.84\%) and the full-modal performance on the server (83.83\%).

\subsection{\color{red}Performance with Varying Numbers of Clients and Degrees of Data Heterogeneity}

{\color{red}
To assess the generalizing capability of our framework, we experiment with different numbers of clients (four or six) and varying degrees of data heterogeneity (two or three modalities per client) on BraTS 2018.
Concretely, we divide all available data into seven distinct, non-overlapping subsets of roughly equal sizes, while ensuring that all cases from the same institute are grouped in the same subset.
Of the subsets, one is designated as the server, and the remaining ones are allocated to up to six clients.
Note that the same subsets are consistently used for the server and clients across different combinations of client and modality numbers, so the performance variations can be directly attributed to the impact of varying client numbers and data heterogeneity.
The results are shown in Table \ref{tab:varying_num}.
We can observe that (1) with two modalities per client, more clients result in overall better performance, and (2) with four clients, more modalities per client lead to overall better performance, too---with improved mDSCs for the server and three of the four comparable clients (and their average) in both comparisons.
In addition, our framework consistently improves the performance of the clients and the server upon the baseline local models, and outperforms all classical and SOTA FL methods being compared---similar to the trends in Table \ref{tabs:state-of-art}.
These results confirm the effectiveness and superiority of our framework over existing approaches with different numbers of clients and varying degrees of data heterogeneity.
}

\input{sections/tabs/tab7.tex}

%% file: sections/tabs/tab_test1.tex
\begin{sidewaystable*}[htp]
\centering
\setlength{\tabcolsep}{0.8mm}
\caption{Experimental results on the test data in mDSC (\%).
*: $p<0.05$ comparing against our method in each column.
}%
\begin{adjustbox}{width=\textwidth}
\begin{tabular}{c|cccccccccc}
\hline
\multicolumn{11}{c}{\textit{BraTS 2018}}\\
\hline
Method &     C & T2                & {F/C}                  & T1/T2                   & F/C/T1                   & F/T1/T2                  & F/C/T1/T2                     &  F/C/T1/T2                & Avg                  & S                         \\ \hline
Local models            & 42.37\scriptsize{$\pm$32.76}&48.13\scriptsize{$\pm$32.27}&87.74\scriptsize{$\pm$10.08}&64.93\scriptsize{$\pm$20.85}&71.59\scriptsize{$\pm$25.39}&63.99\scriptsize{$\pm$24.21}&\underline{89.15}\scriptsize{$\pm$6.76}&67.67\scriptsize{$\pm$24.93}&66.95\scriptsize{$\pm$15.49}$^*$&82.56\scriptsize{$\pm$16.75}      \\ 
RFNet \citep{ding2021rfnet} & 39.86\scriptsize{$\pm$33.38}&50.44\scriptsize{$\pm$34.56}&85.50\scriptsize{$\pm$9.44}&68.14\scriptsize{$\pm$20.19}&68.93\scriptsize{$\pm$32.04}&64.96\scriptsize{$\pm$26.20}&85.36\scriptsize{$\pm$8.56}&45.86\scriptsize{$\pm$34.21}$^*$&65.87\scriptsize{$\pm$16.39}$^*$&83.85\scriptsize{$\pm$14.19} \\
\hline
FedAvg \citep{mcmahan2017communication}          &        18.46\scriptsize{$\pm$23.24}&42.12\scriptsize{$\pm$26.12}&82.11\scriptsize{$\pm$14.55}&59.59\scriptsize{$\pm$23.13}&61.13\scriptsize{$\pm$29.89}&61.91\scriptsize{$\pm$22.18}&84.88\scriptsize{$\pm$16.66}&62.09\scriptsize{$\pm$28.27}$^*$&59.04\scriptsize{$\pm$19.86}$^*$&80.10\scriptsize{$\pm$18.55}$^*$\\
perFL \citep{wang2019federated}                   & 17.55\scriptsize{$\pm$22.01}&44.67\scriptsize{$\pm$30.78}&82.69\scriptsize{$\pm$15.26}&61.16\scriptsize{$\pm$23.11}&63.00\scriptsize{$\pm$27.28}&57.90\scriptsize{$\pm$24.74}&84.40\scriptsize{$\pm$16.47}&61.35\scriptsize{$\pm$28.09}$^*$&59.09\scriptsize{$\pm$19.87}$^*$&79.38\scriptsize{$\pm$20.03}$^*$      \\
FedNorm \citep{bernecker2022fednorm}                & 19.97\scriptsize{$\pm$18.90}&44.66\scriptsize{$\pm$31.07}&84.17\scriptsize{$\pm$14.81}&60.22\scriptsize{$\pm$23.40}&66.59\scriptsize{$\pm$25.85}&59.08\scriptsize{$\pm$24.18}&84.78\scriptsize{$\pm$9.06}&59.29\scriptsize{$\pm$28.50}$^*$&59.97\scriptsize{$\pm$19.95}$^*$&79.63\scriptsize{$\pm$17.99}$^*$       \\
IOP-FL \citep{jiang2023iop}          &23.43\scriptsize{$\pm$24.34}&43.68\scriptsize{$\pm$29.70}&91.76\scriptsize{$\pm$3.78}&52.49\scriptsize{$\pm$29.39}&73.18\scriptsize{$\pm$21.57}&62.96\scriptsize{$\pm$20.80}&81.38\scriptsize{$\pm$18.57}&62.03\scriptsize{$\pm$25.74}$^*$&61.36\scriptsize{$\pm$20.30}$^*$&80.16\scriptsize{$\pm$18.94}$^*$\\  
\color{red}FedCostWAvg \citep{machler2021fedcostwavg}&\color{red}34.63\scriptsize{$\pm$28.74}&\color{red}44.66\scriptsize{$\pm$29.94}&\color{red}90.02\scriptsize{$\pm$6.41}&\color{red}65.11\scriptsize{$\pm$25.02}&\color{red}77.95\scriptsize{$\pm$21.87}&\color{red}67.46\scriptsize{$\pm$19.94}&\color{red}82.08\scriptsize{$\pm$19.86}&\color{red}62.12\scriptsize{$\pm$26.58}$^*$&\color{red}65.50\scriptsize{$\pm$17.44}$^*$&\color{red}81.14\scriptsize{$\pm$18.72}$^*$  \\
\color{red}FedPIDAvg \citep{machler2022fedpidavg}&\color{red}36.79\scriptsize{$\pm$26.39}&\color{red}44.30\scriptsize{$\pm$31.00}&\color{red}\underline{93.40}\scriptsize{$\pm$2.99}&\color{red}64.46\scriptsize{$\pm$25.73}&\color{red}79.20\scriptsize{$\pm$21.88}&\color{red}65.72\scriptsize{$\pm$19.73}&\color{red}81.38\scriptsize{$\pm$18.88}&\color{red}62.39\scriptsize{$\pm$26.24}$^*$&\color{red}65.95\scriptsize{$\pm$17.68}$^*$&\color{red}81.90\scriptsize{$\pm$18.53}$^*$  \\
CreamFL \citep{yu2023multimodal}                  & 34.62\scriptsize{$\pm$26.53}&51.70\scriptsize{$\pm$28.44}&89.42\scriptsize{$\pm$7.00}&66.46\scriptsize{$\pm$24.19}&78.15\scriptsize{$\pm$21.74}&68.94\scriptsize{$\pm$19.32}&82.40\scriptsize{$\pm$17.69}$^*$&65.97\scriptsize{$\pm$26.59}&67.21\scriptsize{$\pm$16.40}$^*$&82.83\scriptsize{$\pm$17.06}$^*$  \\ 
FedIoT \citep{zhao2022multimodal}                 & 41.97\scriptsize{$\pm$26.75}&48.33\scriptsize{$\pm$28.52}&{92.35}\scriptsize{$\pm$3.44}&61.69\scriptsize{$\pm$25.15}&81.81\scriptsize{$\pm$22.34}&\underline{70.66}\scriptsize{$\pm$20.16}&88.31\scriptsize{$\pm$6.33}&\underline{68.36}\scriptsize{$\pm$25.86}&69.18\scriptsize{$\pm$16.94}$^*$&\underline{84.89}\scriptsize{$\pm$13.62}      \\ 
FedMSplit \citep{chen2022fedmsplit}              & \underline{48.99}\scriptsize{$\pm$34.20}&\underline{54.09}\scriptsize{$\pm$28.32}&{92.16}\scriptsize{$\pm$4.04}&\underline{68.21}\scriptsize{$\pm$23.34}&\underline{82.48}\scriptsize{$\pm$22.35}&69.92\scriptsize{$\pm$25.61}&87.87\scriptsize{$\pm$7.98}&66.09\scriptsize{$\pm$25.57}&\underline{71.23}\scriptsize{$\pm$14.44}$^*$&79.93\scriptsize{$\pm$17.81}$^*$        \\
 \hline
Ours   & 
\textbf{58.87}\scriptsize{$\pm$30.19}&\textbf{59.35}\scriptsize{$\pm$31.56}&\textbf{93.73}\scriptsize{$\pm$2.81}&\textbf{75.83}\scriptsize{$\pm$20.36}&\textbf{82.99}\scriptsize{$\pm$22.60}&\textbf{74.58}\scriptsize{$\pm$21.21}&\textbf{90.69}\scriptsize{$\pm$5.44}&\textbf{69.62}\scriptsize{$\pm$26.62}&\textbf{75.70}\scriptsize{$\pm$11.78}&\textbf{84.98}\scriptsize{$\pm$12.73} \\ 
\hline
\end{tabular}
\end{adjustbox}

\begin{adjustbox}{width=\textwidth}
\begin{tabular}{c|cccccccccc}
\hline
\multicolumn{11}{c}{\textit{BraTS 2020}}\\
\hline
Method &     C & T2                & F/C                  & T1/T2                   & F/C/T1                   & F/T1/T2                  & F/C/T1/T2                     &  F/C/T1/T2                & Avg                  & S                                \\ \hline
Local models            & \underline{88.86}\scriptsize{$\pm$7.21}&54.54\scriptsize{$\pm$27.04}$^*$&58.52\scriptsize{$\pm$32.04}&70.03\scriptsize{$\pm$17.52}$^*$&80.93\scriptsize{$\pm$24.42}&61.63\scriptsize{$\pm$27.78}$^*$&88.01\scriptsize{$\pm$6.68}$^*$&68.53\scriptsize{$\pm$29.10}&71.38\scriptsize{$\pm$12.38}$^*$&88.07\scriptsize{$\pm$12.51}$^*$        \\ 
{RFNet} \citep{ding2021rfnet} & 87.11\scriptsize{$\pm$8.95}&\underline{58.75}\scriptsize{$\pm$32.23}&58.51\scriptsize{$\pm$37.48}&69.24\scriptsize{$\pm$17.49}&72.65\scriptsize{$\pm$31.49}&58.09\scriptsize{$\pm$29.11}&87.54\scriptsize{$\pm$6.52}&52.16\scriptsize{$\pm$35.48}$^*$&70.33\scriptsize{$\pm$13.67}$^*$&88.24\scriptsize{$\pm$9.36} \\\hline
FedAvg \citep{mcmahan2017communication}         &         71.34\scriptsize{$\pm$13.97}$^*$&40.86\scriptsize{$\pm$28.40}$^*$&60.68\scriptsize{$\pm$31.47}&55.40\scriptsize{$\pm$24.90}$^*$&66.60\scriptsize{$\pm$33.14}$^*$&57.01\scriptsize{$\pm$26.02}$^*$&83.35\scriptsize{$\pm$15.49}$^*$&60.01\scriptsize{$\pm$30.87}$^*$&61.91\scriptsize{$\pm$11.65}$^*$&87.61\scriptsize{$\pm$12.58}$^*$\\
perFL \citep{wang2019federated}                   & 72.07\scriptsize{$\pm$14.85}$^*$&41.18\scriptsize{$\pm$29.19}$^*$&59.47\scriptsize{$\pm$31.51}&59.89\scriptsize{$\pm$24.01}$^*$&67.99\scriptsize{$\pm$29.58}$^*$&56.55\scriptsize{$\pm$26.23}$^*$&83.78\scriptsize{$\pm$13.70}$^*$&61.98\scriptsize{$\pm$30.06}$^*$&62.86\scriptsize{$\pm$11.62}$^*$&87.40\scriptsize{$\pm$12.51}$^*$      \\
FedNorm \citep{bernecker2022fednorm}                & 76.14\scriptsize{$\pm$11.46}$^*$&40.53\scriptsize{$\pm$26.92}$^*$&57.34\scriptsize{$\pm$31.58}$^*$&56.16\scriptsize{$\pm$31.16}$^*$&68.45\scriptsize{$\pm$27.79}$^*$&57.97\scriptsize{$\pm$27.66}$^*$&84.17\scriptsize{$\pm$13.15}$^*$&63.94\scriptsize{$\pm$31.02}$^*$&63.09\scriptsize{$\pm$15.07}$^*$&86.55\scriptsize{$\pm$13.47}$^*$       \\
IOP-FL \citep{jiang2023iop}                    &76.57\scriptsize{$\pm$17.37}$^*$&53.39\scriptsize{$\pm$25.51}$^*$&58.43\scriptsize{$\pm$31.46}$^*$&64.87\scriptsize{$\pm$15.69}&78.36\scriptsize{$\pm$12.85}$^*$&63.83\scriptsize{$\pm$22.55}&80.91\scriptsize{$\pm$16.98}$^*$&63.39\scriptsize{$\pm$29.10}$^*$&67.47\scriptsize{$\pm$9.34}$^*$&88.19\scriptsize{$\pm$7.81} \\
\color{red}FedCostWAvg \citep{machler2021fedcostwavg} &\color{red} 83.56\scriptsize{$\pm$8.14}$^*$&\color{red}48.46\scriptsize{$\pm$31.19}$^*$&\color{red}64.18\scriptsize{$\pm$32.14}&\color{red}61.98\scriptsize{$\pm$23.31}$^*$&\color{red}73.17\scriptsize{$\pm$29.51}$^*$&\color{red}54.32\scriptsize{$\pm$26.23}$^*$&\color{red}87.56\scriptsize{$\pm$8.45}$^*$&\color{red}65.84\scriptsize{$\pm$27.63}$^*$&\color{red}67.38\scriptsize{$\pm$12.61}$^*$&\color{red}86.05\scriptsize{$\pm$14.36}$^*$ \\
\color{red}FedPIDAvg \citep{machler2022fedpidavg} &\color{red} 83.78\scriptsize{$\pm$8.73}$^*$&\color{red}48.30\scriptsize{$\pm$32.05}$^*$&\color{red}\underline{64.73}\scriptsize{$\pm$29.26}&\color{red}63.38\scriptsize{$\pm$23.76}$^*$&\color{red}75.73\scriptsize{$\pm$28.41}$^*$&\color{red}49.83\scriptsize{$\pm$26.70}$^*$&\color{red}88.13\scriptsize{$\pm$7.57}$^*$&\color{red}67.22\scriptsize{$\pm$28.19}$^*$&\color{red}67.64\scriptsize{$\pm$13.53}$^*$&\color{red}85.98\scriptsize{$\pm$13.74}$^*$\\
CreamFL \citep{yu2023multimodal}                  & 85.43\scriptsize{$\pm$9.00}$^*$&49.76\scriptsize{$\pm$28.35}$^*$&54.99\scriptsize{$\pm$31.88}$^*$&60.19\scriptsize{$\pm$24.22}$^*$&78.90\scriptsize{$\pm$24.09}$^*$&56.70\scriptsize{$\pm$26.99}$^*$&85.33\scriptsize{$\pm$11.78}$^*$&65.42\scriptsize{$\pm$29.36}$^*$&67.09\scriptsize{$\pm$13.29}$^*$&87.69\scriptsize{$\pm$11.65}$^*$        \\ 
FedIoT \citep{zhao2022multimodal}                & 86.97\scriptsize{$\pm$8.79}$^*$&53.92\scriptsize{$\pm$29.15}$^*$&58.64\scriptsize{$\pm$31.96}$^*$&70.87\scriptsize{$\pm$12.31}$^*$&79.78\scriptsize{$\pm$24.99}$^*$&65.64\scriptsize{$\pm$22.14}&88.96\scriptsize{$\pm$4.45}$^*$&64.79\scriptsize{$\pm$29.69}$^*$&71.20\scriptsize{$\pm$12.07}$^*$&\underline{88.77}\scriptsize{$\pm$9.70}        \\
FedMSplit \citep{chen2022fedmsplit}              & 87.53\scriptsize{$\pm$8.73}$^*$&56.28\scriptsize{$\pm$29.58}$^*$&{63.03}\scriptsize{$\pm$34.91}&\underline{76.71}\scriptsize{$\pm$11.93}&\underline{81.32}\scriptsize{$\pm$24.34}&\underline{67.46}\scriptsize{$\pm$24.33}&\underline{89.09}\scriptsize{$\pm$6.51}&\underline{68.97}\scriptsize{$\pm$28.83}$^*$&\underline{73.80}\scriptsize{$\pm$11.03}$^*$&86.88\scriptsize{$\pm$13.54}$^*$                        \\
 \hline
Ours   & \textbf{89.21}\scriptsize{$\pm$7.59}&\textbf{62.68}\scriptsize{$\pm$29.12}&\textbf{65.25}\scriptsize{$\pm$31.87}&\textbf{76.74}\scriptsize{$\pm$13.93}&\textbf{81.93}\scriptsize{$\pm$24.77}&\textbf{69.91}\scriptsize{$\pm$22.13}&\textbf{90.21}\scriptsize{$\pm$5.10}&\textbf{71.28}\scriptsize{$\pm$28.87}&\textbf{75.90}\scriptsize{$\pm$9.78}&\textbf{89.39}\scriptsize{$\pm$8.78}      \\ \hline
\end{tabular}
\end{adjustbox}
\label{tabs:state-of-art}

\end{sidewaystable*}

%% file: sections/tabs/tab9_hd95.tex
\begin{sidewaystable*}[htp]\color{red}
\centering
\setlength{\tabcolsep}{0.8mm}
\caption{Experimental results on the test data in HD95 (pixel).
*: $p<0.05$ comparing against our method in each column.
}%
\begin{adjustbox}{width=\textwidth}
\begin{tabular}{c|cccccccccc}
\hline
\multicolumn{11}{c}{\textit{BraTS 2018}}\\
\hline
Method &     C & T2                & {F/C}                  & T1/T2                   & F/C/T1                   & F/T1/T2                  & F/C/T1/T2                     &  F/C/T1/T2                & Avg                  & S                         \\ \hline
Local models            & 58.50\scriptsize{$\pm$33.37}&17.40\scriptsize{$\pm$10.75}&29.82\scriptsize{$\pm$32.46}&25.97\scriptsize{$\pm$27.22}&10.40\scriptsize{$\pm$19.64}&21.10\scriptsize{$\pm$19.98}&22.08\scriptsize{$\pm$31.75}&24.60\scriptsize{$\pm$26.17}&26.23\scriptsize{$\pm$13.36}$^*$&11.59\scriptsize{$\pm$24.93}      \\ 
RFNet \citep{ding2021rfnet} & 51.18\scriptsize{$\pm$28.54}&\underline{18.89}\scriptsize{$\pm$10.86}&6.01\scriptsize{$\pm$3.24}&\underline{12.58}\scriptsize{$\pm$7.15}&6.23\scriptsize{$\pm$3.64}&10.66\scriptsize{$\pm$7.23}&21.91\scriptsize{$\pm$35.08}&19.48\scriptsize{$\pm$16.65}&{18.37}\scriptsize{$\pm$12.80}&12.11\scriptsize{$\pm$24.52} \\
\hline
FedAvg \citep{mcmahan2017communication}          &        62.12\scriptsize{$\pm$19.56}&38.87\scriptsize{$\pm$36.99}&4.44\scriptsize{$\pm$3.74}&28.55\scriptsize{$\pm$30.32}&8.30\scriptsize{$\pm$5.27}&12.16\scriptsize{$\pm$13.48}&16.50\scriptsize{$\pm$33.10}&16.52\scriptsize{$\pm$12.73}&23.43\scriptsize{$\pm$17.94}$^*$&14.52\scriptsize{$\pm$27.38}\\
perFL \citep{wang2019federated}                   & 67.95\scriptsize{$\pm$20.95}&39.88\scriptsize{$\pm$16.85}&17.85\scriptsize{$\pm$25.37}&25.99\scriptsize{$\pm$29.84}&13.55\scriptsize{$\pm$15.25}&8.30\scriptsize{$\pm$3.50}&25.28\scriptsize{$\pm$36.74}&\underline{14.54}\scriptsize{$\pm$12.98}&26.67\scriptsize{$\pm$18.08}$^*$&12.64\scriptsize{$\pm$25.51}      \\
FedNorm \citep{bernecker2022fednorm}                & 65.52\scriptsize{$\pm$16.42}&24.87\scriptsize{$\pm$8.32}&3.35\scriptsize{$\pm$3.24}&17.37\scriptsize{$\pm$14.80}&13.65\scriptsize{$\pm$16.26}&19.26\scriptsize{$\pm$20.99}&28.02\scriptsize{$\pm$35.74}$^*$&29.83\scriptsize{$\pm$28.19}&25.23\scriptsize{$\pm$17.19}$^*$&15.13\scriptsize{$\pm$27.07}$^*$    \\
IOP-FL \citep{jiang2023iop}          &69.30\scriptsize{$\pm$23.61}&23.40\scriptsize{$\pm$8.50}&5.72\scriptsize{$\pm$10.62}&28.36\scriptsize{$\pm$26.32}&24.16\scriptsize{$\pm$23.62}&8.10\scriptsize{$\pm$3.60}&17.24\scriptsize{$\pm$30.80}&17.93\scriptsize{$\pm$19.40}&24.28\scriptsize{$\pm$18.51}$^*$&10.51\scriptsize{$\pm$23.32}\\  
\color{red}FedCostWAvg \citep{machler2021fedcostwavg}                 & 78.82\scriptsize{$\pm$28.56}&19.26\scriptsize{$\pm$10.57}&14.26\scriptsize{$\pm$26.75}&28.96\scriptsize{$\pm$31.93}&14.54\scriptsize{$\pm$18.26}&9.46\scriptsize{$\pm$6.27}&\underline{15.77}\scriptsize{$\pm$33.08}&22.41\scriptsize{$\pm$25.04}&25.43\scriptsize{$\pm$20.92}$^*$&13.04\scriptsize{$\pm$27.75} \\ 
\color{red}FedPIDAvg \citep{machler2022fedpidavg}                 & 73.92\scriptsize{$\pm$27.86}&20.52\scriptsize{$\pm$10.26}&\underline{2.23}\scriptsize{$\pm$1.54}&25.73\scriptsize{$\pm$31.57}&5.99\scriptsize{$\pm$4.19}&29.12\scriptsize{$\pm$31.32}&22.42\scriptsize{$\pm$33.59}&26.68\scriptsize{$\pm$30.92}&25.83\scriptsize{$\pm$20.35}$^*$&12.33\scriptsize{$\pm$27.15}    \\ 
CreamFL \citep{yu2023multimodal}              & \underline{49.20}\scriptsize{$\pm$27.62}&28.51\scriptsize{$\pm$19.36}&18.46\scriptsize{$\pm$28.39}&37.14\scriptsize{$\pm$35.07}&5.10\scriptsize{$\pm$2.97}&13.39\scriptsize{$\pm$13.28}&15.81\scriptsize{$\pm$32.47}&25.53\scriptsize{$\pm$32.32}&24.14\scriptsize{$\pm$13.22}$^*$&11.31\scriptsize{$\pm$24.34}     \\
FedIoT \citep{zhao2022multimodal} & 72.32\scriptsize{$\pm$32.83}&22.02\scriptsize{$\pm$13.60}&19.91\scriptsize{$\pm$23.89}&27.29\scriptsize{$\pm$28.22}&4.01\scriptsize{$\pm$2.56}&10.22\scriptsize{$\pm$10.19}&21.00\scriptsize{$\pm$34.10}&22.24\scriptsize{$\pm$18.44}&24.88\scriptsize{$\pm$19.24}$^*$&\underline{9.66}\scriptsize{$\pm$22.37} \\
FedMSplit \citep{chen2022fedmsplit} & \textbf{36.68}\scriptsize{$\pm$29.18}&20.11\scriptsize{$\pm$12.76}&3.26\scriptsize{$\pm$3.42}&23.92\scriptsize{$\pm$30.78}&\underline{3.66}\scriptsize{$\pm$1.74}&\underline{6.83}\scriptsize{$\pm$3.62}&25.33\scriptsize{$\pm$32.16}&24.32\scriptsize{$\pm$27.24}&\underline{18.01}\scriptsize{$\pm$11.34}&12.40\scriptsize{$\pm$24.16} \\



 \hline
Ours   & 
{51.16}\scriptsize{$\pm$34.14}&\textbf{14.11}\scriptsize{$\pm$10.29}&\textbf{2.14}\scriptsize{$\pm$1.36}&\textbf{7.29}\scriptsize{$\pm$4.32}&\textbf{3.32}\scriptsize{$\pm$1.85}&\textbf{5.12}\scriptsize{$\pm$2.38}&\textbf{9.73}\scriptsize{$\pm$20.92}&\textbf{10.99}\scriptsize{$\pm$8.52}&\textbf{12.98}\scriptsize{$\pm$14.91}$^*$&\textbf{6.52}\scriptsize{$\pm$9.42} \\ 
\hline
\end{tabular}
\end{adjustbox}

\begin{adjustbox}{width=\textwidth}
\begin{tabular}{c|cccccccccc}
\hline
\multicolumn{11}{c}{\textit{BraTS 2020}}\\
\hline
Method &     C & T2                & F/C                  & T1/T2                   & F/C/T1                   & F/T1/T2                  & F/C/T1/T2                     &  F/C/T1/T2                & Avg                  & S                                \\ \hline
Local models            & 6.10\scriptsize{$\pm$8.50}$^*$&37.12\scriptsize{$\pm$34.45}$^*$&24.27\scriptsize{$\pm$20.71}$^*$&13.09\scriptsize{$\pm$14.57}&9.95\scriptsize{$\pm$17.01}&29.08\scriptsize{$\pm$27.21}$^*$&8.65\scriptsize{$\pm$13.88}&25.18\scriptsize{$\pm$32.77}&19.18\scriptsize{$\pm$10.52}$^*$&6.83\scriptsize{$\pm$18.17}        \\ 
{RFNet} \citep{ding2021rfnet} & 7.01\scriptsize{$\pm$7.99}&\underline{24.25}\scriptsize{$\pm$22.47}&23.15\scriptsize{$\pm$21.64}&15.52\scriptsize{$\pm$12.32}&8.42\scriptsize{$\pm$10.37}&28.04\scriptsize{$\pm$29.95}&8.69\scriptsize{$\pm$11.28}&28.79\scriptsize{$\pm$33.11}$^*$&17.98\scriptsize{$\pm$9.43}$^*$&6.47\scriptsize{$\pm$18.46} \\\hline
FedAvg \citep{mcmahan2017communication}         &         36.25\scriptsize{$\pm$20.31}$^*$&41.37\scriptsize{$\pm$30.47}$^*$&18.47\scriptsize{$\pm$12.91}&25.76\scriptsize{$\pm$17.37}&17.52\scriptsize{$\pm$19.77}$^*$&15.32\scriptsize{$\pm$10.55}$^*$&10.04\scriptsize{$\pm$15.12}&28.49\scriptsize{$\pm$36.51}&24.15\scriptsize{$\pm$10.11}$^*$&8.11\scriptsize{$\pm$19.93}$^*$\\
perFL \citep{wang2019federated}                   & 22.64\scriptsize{$\pm$21.46}$^*$&30.58\scriptsize{$\pm$28.27}$^*$&18.26\scriptsize{$\pm$13.86}&20.57\scriptsize{$\pm$13.97}&17.08\scriptsize{$\pm$19.79}$^*$&18.51\scriptsize{$\pm$15.68}&\textbf{5.81}\scriptsize{$\pm$8.08}&27.00\scriptsize{$\pm$35.17}&20.05\scriptsize{$\pm$6.92}$^*$&8.84\scriptsize{$\pm$20.67}$^*$     \\
FedNorm \citep{bernecker2022fednorm}                & 25.76\scriptsize{$\pm$17.98}$^*$&39.75\scriptsize{$\pm$27.90}$^*$&30.78\scriptsize{$\pm$23.05}$^*$&47.54\scriptsize{$\pm$30.42}&29.87\scriptsize{$\pm$36.65}$^*$&20.45\scriptsize{$\pm$14.82}$^*$&7.93\scriptsize{$\pm$12.90}&24.94\scriptsize{$\pm$29.94}&28.38\scriptsize{$\pm$11.20}$^*$&9.64\scriptsize{$\pm$23.02}$^*$       \\
IOP-FL \citep{jiang2023iop}                    &32.58\scriptsize{$\pm$40.28}$^*$&20.95\scriptsize{$\pm$16.58}$^*$&19.59\scriptsize{$\pm$21.23}$^*$&19.78\scriptsize{$\pm$17.74}&20.97\scriptsize{$\pm$36.44}&9.82\scriptsize{$\pm$3.88}&19.31\scriptsize{$\pm$24.56}&\textbf{7.11}\scriptsize{$\pm$4.77}&18.76\scriptsize{$\pm$7.23}$^*$&6.13\scriptsize{$\pm$16.79}$^*$ \\
\color{red}FedCostWAvg \citep{machler2021fedcostwavg} & 6.90\scriptsize{$\pm$7.04}$^*$&21.30\scriptsize{$\pm$23.37}&14.11\scriptsize{$\pm$13.15}&19.00\scriptsize{$\pm$11.56}&8.20\scriptsize{$\pm$8.30}$^*$&14.13\scriptsize{$\pm$11.80}$^*$&9.40\scriptsize{$\pm$17.75}&27.47\scriptsize{$\pm$32.96}$^*$&15.06\scriptsize{$\pm$6.67}$^*$&9.01\scriptsize{$\pm$22.45} \\
\color{red}FedPIDAvg \citep{machler2022fedpidavg} & 6.58\scriptsize{$\pm$5.97}$^*$&27.40\scriptsize{$\pm$30.56}&18.00\scriptsize{$\pm$18.58}&17.25\scriptsize{$\pm$15.72}&\underline{5.86}\scriptsize{$\pm$5.93}$^*$&26.83\scriptsize{$\pm$25.78}$^*$&8.18\scriptsize{$\pm$13.49}&18.11\scriptsize{$\pm$25.56}&16.03\scriptsize{$\pm$7.98}$^*$&10.65\scriptsize{$\pm$23.16}$^*$\\
CreamFL \citep{yu2023multimodal}                  & 14.40\scriptsize{$\pm$19.61}$^*$&38.52\scriptsize{$\pm$37.81}$^*$&22.76\scriptsize{$\pm$18.30}&15.36\scriptsize{$\pm$10.91}&13.18\scriptsize{$\pm$25.78}$^*$&15.04\scriptsize{$\pm$10.41}$^*$&9.67\scriptsize{$\pm$15.82}&26.11\scriptsize{$\pm$31.15}&19.38\scriptsize{$\pm$8.77}$^*$&8.90\scriptsize{$\pm$20.66}$^*$        \\ 
FedIoT \citep{zhao2022multimodal}                & 16.01\scriptsize{$\pm$27.35}$^*$&21.00\scriptsize{$\pm$20.92}$^*$&\underline{13.90}\scriptsize{$\pm$13.17}&22.64\scriptsize{$\pm$28.51}&10.41\scriptsize{$\pm$18.56}&\underline{8.43}\scriptsize{$\pm$2.87}&12.78\scriptsize{$\pm$18.97}$^*$&15.75\scriptsize{$\pm$22.07}$^*$&15.12\scriptsize{$\pm$4.56}$^*$&\underline{3.48}\scriptsize{$\pm$4.33}        \\
FedMSplit \citep{chen2022fedmsplit}              & \underline{4.34}\scriptsize{$\pm$3.15}$^*$&\underline{15.34}\scriptsize{$\pm$18.18}&23.25\scriptsize{$\pm$22.79}&\underline{10.85}\scriptsize{$\pm$6.98}&23.23\scriptsize{$\pm$30.82}$^*$&15.18\scriptsize{$\pm$12.34}&8.07\scriptsize{$\pm$13.45}&17.03\scriptsize{$\pm$25.19}$^*$&\underline{14.66}\scriptsize{$\pm$6.31}$^*$&9.14\scriptsize{$\pm$21.21}$^*$                  \\

 \hline
Ours   & \textbf{3.82}\scriptsize{$\pm$2.59}$^*$&\textbf{11.06}\scriptsize{$\pm$10.21}$^*$&\textbf{12.55}\scriptsize{$\pm$14.30}$^*$&\textbf{6.46}\scriptsize{$\pm$2.99}&\textbf{2.84}\scriptsize{$\pm$1.87}&\textbf{8.19}\scriptsize{$\pm$3.26}$^*$&\underline{8.03}\scriptsize{$\pm$13.76}&\underline{13.35}\scriptsize{$\pm$22.58}&\textbf{8.29}\scriptsize{$\pm$3.62}$^*$&\textbf{2.94}\scriptsize{$\pm$3.95}      \\ \hline
\end{tabular}
\end{adjustbox}
\label{tabs:state-of-art_hd}

\end{sidewaystable*}

%% file: sections/tabs/tab5.tex
\begin{table*}[t]
\centering
\setlength{\tabcolsep}{.8mm}
\caption{Experimental resutls on the \textit{validation} data of BraTS 2018 in mDSC (\%).
Top: varying $N_k$ (number of multimodal anchors per class) with $l=4$. 
Bottom: varying the feature scale level $l$ (with $N_k=4$) of the multimodal fusion decoder $D_M$, based on which the cluster membership is determined;
$l=4$ indicates the most abstract level of the smallest scale (i.e., at the bottleneck between the encoders and decoder), and ``1--4'' concatenates features of all four levels together for clustering.
*: $p<0.05$ comparing against $N_k=4$ (top) and $l=4$ (bottom), respectively, in each column.}%
\begin{adjustbox}{width=.99\textwidth}
\begin{tabular}{c|cccccccccc}
\hline
$N_k$ & C & T2                & F/C                  & T1/T2                   & F/C/T1                   & F/T1/T2                  & F/C/T1/T2                     &  F/C/T1/T2                & Avg                             & S                 \\ \hline
1                 & 48.32\scriptsize{$\pm$31.53}&61.34\scriptsize{$\pm$27.33}&85.19\scriptsize{$\pm$11.33}&75.48\scriptsize{$\pm$14.15}&67.37\scriptsize{$\pm$31.41}&70.52\scriptsize{$\pm$25.21}&90.67\scriptsize{$\pm$5.43}&70.64\scriptsize{$\pm$25.89}&71.19\scriptsize{$\pm$12.40}&83.71\scriptsize{$\pm$15.89}\\
2                 & 50.27\scriptsize{$\pm$30.28}&\textbf{63.30}\scriptsize{$\pm$25.20}&84.22\scriptsize{$\pm$12.49}&74.31\scriptsize{$\pm$14.32}&67.24\scriptsize{$\pm$31.47}&74.72\scriptsize{$\pm$16.01}&{90.87}\scriptsize{$\pm$5.56}$^*$&70.36\scriptsize{$\pm$26.76}&71.91\scriptsize{$\pm$11.68}&83.56\scriptsize{$\pm$16.13} \\
3       & 44.54\scriptsize{$\pm$35.29}&59.83\scriptsize{$\pm$26.82}&\textbf{86.59}\scriptsize{$\pm$9.54}&74.16\scriptsize{$\pm$14.87}&\textbf{75.63}\scriptsize{$\pm$24.81}&\textbf{76.06}\scriptsize{$\pm$15.66}&90.29\scriptsize{$\pm$5.94}&69.91\scriptsize{$\pm$26.34}&72.12\scriptsize{$\pm$13.64}&83.51\scriptsize{$\pm$16.06}\\
\rowcolor[HTML]{EFEFEF}4                 & \textbf{51.04}\scriptsize{$\pm$29.58}&{62.81}\scriptsize{$\pm$27.02}&84.76\scriptsize{$\pm$10.21}&\textbf{74.27}\scriptsize{$\pm$13.20}&{72.84}\scriptsize{$\pm$26.11}&{73.87}\scriptsize{$\pm$20.79}&\textbf{91.02}\scriptsize{$\pm$5.28}&\textbf{72.08}\scriptsize{$\pm$24.90}&\textbf{72.84}\scriptsize{$\pm$11.47}&\textbf{83.83}\scriptsize{$\pm$15.43} \\
5                 &

45.39\scriptsize{$\pm$35.60}&58.20\scriptsize{$\pm$30.65}&86.53\scriptsize{$\pm$5.38}&73.38\scriptsize{$\pm$14.31}&74.72\scriptsize{$\pm$25.79}&75.29\scriptsize{$\pm$13.27}&90.43\scriptsize{$\pm$5.82}&{70.80}\scriptsize{$\pm$26.43}&71.84\scriptsize{$\pm$13.56}&83.74\scriptsize{$\pm$15.10}\\ 
6                 &49.59\scriptsize{$\pm$29.97}&57.64\scriptsize{$\pm$28.77}&85.00\scriptsize{$\pm$9.50}&74.17\scriptsize{$\pm$15.25}&70.43\scriptsize{$\pm$27.55}&74.32\scriptsize{$\pm$17.00}&90.13\scriptsize{$\pm$6.03}&69.53\scriptsize{$\pm$26.43}&71.33\scriptsize{$\pm$12.36}&83.05\scriptsize{$\pm$17.02}\\
\hline

\end{tabular}
\end{adjustbox}

\vspace{.5mm}

\begin{adjustbox}{width=.99\textwidth}
\begin{tabular}{c|cccccccccc}
\hline
{\phantom{ii}$l$\phantom{iii}} & C & T2                & F/C                  & T1/T2                   & F/C/T1                   & F/T1/T2                  & F/C/T1/T2                     &  F/C/T1/T2                & Avg                             & S          \\ \hline
{1}                         & 48.39\scriptsize{$\pm$34.57}&60.44\scriptsize{$\pm$25.84}&84.89\scriptsize{$\pm$8.60}&74.05\scriptsize{$\pm$14.47}&68.74\scriptsize{$\pm$30.98}&74.52\scriptsize{$\pm$16.64}&{90.98}\scriptsize{$\pm$5.31}&70.22\scriptsize{$\pm$26.27}&71.51\scriptsize{$\pm$12.47}&83.93\scriptsize{$\pm$14.31} \\
{2}                         & \textbf{52.15}\scriptsize{$\pm$30.81}&58.73\scriptsize{$\pm$29.07}&83.42\scriptsize{$\pm$12.00}&72.67\scriptsize{$\pm$15.73}&66.28\scriptsize{$\pm$31.61}&72.89\scriptsize{$\pm$18.29}&90.71\scriptsize{$\pm$5.60}&68.87\scriptsize{$\pm$27.05}&70.72\scriptsize{$\pm$11.63}$^*$&84.22\scriptsize{$\pm$14.24} \\
{3}                         & 51.65\scriptsize{$\pm$30.44}&61.94\scriptsize{$\pm$26.28}&\textbf{86.92}\scriptsize{$\pm$7.95}&73.75\scriptsize{$\pm$14.19}&70.03\scriptsize{$\pm$25.10}&73.37\scriptsize{$\pm$19.83}&90.37\scriptsize{$\pm$5.74}&\textbf{72.48}\scriptsize{$\pm$26.66}&72.56\scriptsize{$\pm$11.60}&83.55\scriptsize{$\pm$16.33} \\
\rowcolor[HTML]{EFEFEF}4                  & {51.04}\scriptsize{$\pm$29.58}&\textbf{62.81}\scriptsize{$\pm$27.02}&84.76\scriptsize{$\pm$10.21}&{74.27}\scriptsize{$\pm$13.20}&\textbf{72.84}\scriptsize{$\pm$26.11}&{73.87}\scriptsize{$\pm$20.79}&\textbf{91.02}\scriptsize{$\pm$5.28}&{72.08}\scriptsize{$\pm$24.90}&\textbf{72.84}\scriptsize{$\pm$11.47}&\textbf{83.83}\scriptsize{$\pm$15.43} \\
{1--4}                      & 48.29\scriptsize{$\pm$37.35}&61.87\scriptsize{$\pm$26.92}&85.42\scriptsize{$\pm$11.46}&\textbf{75.27}\scriptsize{$\pm$14.80}&{71.66}\scriptsize{$\pm$25.53}&\textbf{76.33}\scriptsize{$\pm$13.53}&90.57\scriptsize{$\pm$5.39}&70.01\scriptsize{$\pm$27.29}&72.43\scriptsize{$\pm$12.40}&82.91\scriptsize{$\pm$17.95} \\ \hline
\end{tabular}
\end{adjustbox}
\label{tabs:clusNum}\vspace{-5pt}
\end{table*}

%% file: sections/tabs/tab6.tex
\begin{table*}[t]\color{blue}
\centering
\setlength{\tabcolsep}{.8mm}
\caption{Experiment results on the \textit{validation} data of BraTS 2018 in mDSC (\%) by varying $P$ (personalization patience) with $l=4, N_k=4$.
$P=0$ means complete personalization without federating any parameters of the fusion decoder.
*: $p<0.05$ comparing against $P=10$ in each column.}%
\begin{adjustbox}{width=\textwidth}
\begin{tabular}{c|cccccccccc}
\hline
Patience ($P$) & C & T2                & F/C                  & T1/T2                   & F/C/T1                   & F/T1/T2                  & F/C/T1/T2                     &  F/C/T1/T2                & Avg                             & S                 \\ \hline
0                & 44.89\scriptsize{$\pm$34.69}&56.45\scriptsize{$\pm$27.58}&82.68\scriptsize{$\pm$10.50}&70.20\scriptsize{$\pm$16.92}&69.87\scriptsize{$\pm$25.45}&69.26\scriptsize{$\pm$22.57}&89.18\scriptsize{$\pm$6.80}&67.10\scriptsize{$\pm$26.99}$^*$&68.70\scriptsize{$\pm$12.93}$^*$&82.72\scriptsize{$\pm$18.13}\\
6                 & 48.18\scriptsize{$\pm$32.72}&59.09\scriptsize{$\pm$27.84}&\textbf{86.35}\scriptsize{$\pm$11.25}&72.30\scriptsize{$\pm$16.28}&\textbf{75.45}\scriptsize{$\pm$25.40}&74.40\scriptsize{$\pm$18.11}&91.01\scriptsize{$\pm$5.26}&71.57\scriptsize{$\pm$26.73}&72.31\scriptsize{$\pm$12.86}&83.54\scriptsize{$\pm$13.89} \\
8       & 48.48\scriptsize{$\pm$33.10}&61.79\scriptsize{$\pm$26.17}&85.48\scriptsize{$\pm$10.89}&73.26\scriptsize{$\pm$14.60}&73.13\scriptsize{$\pm$27.04}&\textbf{75.67}\scriptsize{$\pm$15.63}&90.41\scriptsize{$\pm$5.73}&69.35\scriptsize{$\pm$27.68}&72.20\scriptsize{$\pm$12.25}&83.76\scriptsize{$\pm$15.32} \\
\rowcolor[HTML]{EFEFEF}10                 & {51.04}\scriptsize{$\pm$29.58}&\textbf{62.81}\scriptsize{$\pm$27.02}&84.76\scriptsize{$\pm$10.21}&\textbf{74.27}\scriptsize{$\pm$13.20}&{72.84}\scriptsize{$\pm$26.11}&{73.87}\scriptsize{$\pm$20.79}&\textbf{91.02}\scriptsize{$\pm$5.28}&\textbf{72.08}\scriptsize{$\pm$24.90}&\textbf{72.84}\scriptsize{$\pm$11.47}&\textbf{83.83}\scriptsize{$\pm$15.43}  \\
12                 & 48.99\scriptsize{$\pm$29.76}&58.74\scriptsize{$\pm$28.26}&85.57\scriptsize{$\pm$9.62}&73.76\scriptsize{$\pm$15.46}&71.90\scriptsize{$\pm$31.23}&72.54\scriptsize{$\pm$20.45}&90.40\scriptsize{$\pm$5.74}&70.50\scriptsize{$\pm$27.06}&71.55\scriptsize{$\pm$12.61}$^*$&83.78\scriptsize{$\pm$15.60}\\
14                 & \textbf{52.28}\scriptsize{$\pm$30.88}&59.64\scriptsize{$\pm$28.11}&84.35\scriptsize{$\pm$9.66}&73.86\scriptsize{$\pm$16.01}&74.05\scriptsize{$\pm$26.26}&73.42\scriptsize{$\pm$19.28}&90.93\scriptsize{$\pm$5.34}&69.77\scriptsize{$\pm$26.64}&72.29\scriptsize{$\pm$11.55}&83.74\scriptsize{$\pm$15.02} \\ \hline
\end{tabular}
\end{adjustbox}\vspace{-4pt}


\label{tabs:patience}
\end{table*}

%% file: sections/tabs/tab8_limit_server.tex
\begin{table*}[t]\color{red}
\centering
\setlength{\tabcolsep}{0.8mm}
\caption{Experimental results on the test data of BraTS 2018 in mDSC (\%), with varying amounts and quality of server data.
FedMSplit \citep{chen2022fedmsplit}, the second-best method in Table \ref{tabs:state-of-art} in terms of average client mDSC, is also shown for reference.
{*: $p<0.05$ comparing against our method with all server data in each column.}
}%
\begin{adjustbox}{width=\textwidth}
\begin{tabular}{c|cccccccccc}
\hline
Method &     C & T2                & {F/C}                  & T1/T2                   & F/C/T1                   & F/T1/T2                  & F/C/T1/T2                     &  F/C/T1/T2                & Avg                  & S                         \\ \hline
Ours (all server data)   &
{58.87}\scriptsize{$\pm$30.19}&{59.35}\scriptsize{$\pm$31.56}&{93.73}\scriptsize{$\pm$2.81}&{75.83}\scriptsize{$\pm$20.36}&{82.99}\scriptsize{$\pm$22.60}&{74.58}\scriptsize{$\pm$21.21}&{90.69}\scriptsize{$\pm$5.44}&{69.62}\scriptsize{$\pm$26.62}&{75.70}\scriptsize{$\pm$11.78}&{84.98}\scriptsize{$\pm$12.73} \\ 
Ours (50\% server data) & 56.53\scriptsize{$\pm$26.34}&59.36\scriptsize{$\pm$31.69}&92.74\scriptsize{$\pm$4.64}&71.24\scriptsize{$\pm$24.43}&81.92\scriptsize{$\pm$23.12}&71.62\scriptsize{$\pm$26.04}&91.36\scriptsize{$\pm$5.36}&69.93\scriptsize{$\pm$27.50}&74.34\scriptsize{$\pm$12.57}&82.98\scriptsize{$\pm$15.45} \\
Ours (30\% server data) & 57.27\scriptsize{$\pm$24.08}&55.91\scriptsize{$\pm$31.02}&92.65\scriptsize{$\pm$4.08}&69.96\scriptsize{$\pm$22.40}&82.69\scriptsize{$\pm$22.67}&72.92\scriptsize{$\pm$21.84}&88.82\scriptsize{$\pm$5.48}&70.29\scriptsize{$\pm$27.12}&73.81\scriptsize{$\pm$12.66}$^*$&80.68\scriptsize{$\pm$16.92}$^*$ \\
Ours (10\% server data) & 50.14\scriptsize{$\pm$26.43}&56.85\scriptsize{$\pm$29.32}&93.39\scriptsize{$\pm$3.70}&69.49\scriptsize{$\pm$23.50}&81.84\scriptsize{$\pm$22.50}&72.10\scriptsize{$\pm$22.07}&89.87\scriptsize{$\pm$5.57}&68.81\scriptsize{$\pm$26.18}&72.81\scriptsize{$\pm$14.11}$^*$&78.30\scriptsize{$\pm$17.12}$^*$ \\
\color{red} Ours (corrupted server data) & 
59.50\scriptsize{$\pm$22.96}&61.61\scriptsize{$\pm$29.26}&93.13\scriptsize{$\pm$3.74}&72.85\scriptsize{$\pm$22.44}&82.23\scriptsize{$\pm$22.52}&70.41\scriptsize{$\pm$26.86}&90.70\scriptsize{$\pm$6.17}&69.70\scriptsize{$\pm$26.01}&75.02\scriptsize{$\pm$11.73}&81.43\scriptsize{$\pm$16.13}$^*$\\
\hline
FedMSplit (all server data) & {48.99}\scriptsize{$\pm$34.20}&{54.09}\scriptsize{$\pm$28.32}&{92.16}\scriptsize{$\pm$4.04}&{68.21}\scriptsize{$\pm$23.34}&{82.48}\scriptsize{$\pm$22.35}&69.92\scriptsize{$\pm$25.61}&87.87\scriptsize{$\pm$7.98}&66.09\scriptsize{$\pm$25.57}&{71.23}\scriptsize{$\pm$14.44}$^*$&79.93\scriptsize{$\pm$17.81}$^*$        \\
\hline
\end{tabular}
\end{adjustbox}
\label{tabs:server_limit}

\end{table*}

%% file: sections/tabs/tab_valid.tex
\begin{table*}[t]
\centering
\setlength{\tabcolsep}{0.8mm}
\caption{Ablation study results on the \textit{validation} data of BraTS 2018 in mDSC (\%).
``E'', ``D'', and ``$\text{D}_{par}$'' stand for the encoder, fully federated decoder, and partially federated (partially personalized) decoder, respectively.
*: $p<0.05$ comparing against our method in each column.}%
\begin{adjustbox}{width=\textwidth}
\begin{tabular}{lccc|cccccccccc}
\hline
Ablation & Server & Federated & LACCA        &   C & T2                & F/C                  & T1/T2                   & F/C/T1                   & F/T1/T2                  & F/C/T1/T2                     &  F/C/T1/T2                & Avg                             & S              \\ \hline
(a)      & E\&D   & E         & -            & 32.50\scriptsize{$\pm$27.50}&41.87\scriptsize{$\pm$27.02}&60.52\scriptsize{$\pm$20.68}&54.75\scriptsize{$\pm$21.71}&38.74\scriptsize{$\pm$26.36}&62.27\scriptsize{$\pm$23.62}&90.31\scriptsize{$\pm$5.34}&61.96\scriptsize{$\pm$25.03}&55.37\scriptsize{$\pm$17.03}$^*$&82.60\scriptsize{$\pm$12.72}$^*$         \\ 
(b)      & E\&D   & D         & -            & 41.49\scriptsize{$\pm$26.56}&52.41\scriptsize{$\pm$31.85}&81.66\scriptsize{$\pm$9.66}&64.33\scriptsize{$\pm$23.47}&56.00\scriptsize{$\pm$30.07}&65.53\scriptsize{$\pm$24.51}&90.03\scriptsize{$\pm$6.16}&66.86\scriptsize{$\pm$27.60}&64.79\scriptsize{$\pm$14.58}$^*$&82.46\scriptsize{$\pm$12.99}$^*$          \\
(c)  & 4E\&D  & 4E        & -            & 44.89\scriptsize{$\pm$34.69}&56.45\scriptsize{$\pm$27.58}&82.68\scriptsize{$\pm$10.50}&70.20\scriptsize{$\pm$16.92}&\underline{69.87}\scriptsize{$\pm$25.45}&69.26\scriptsize{$\pm$22.57}&89.18\scriptsize{$\pm$6.80}&67.10\scriptsize{$\pm$26.99}&68.70\scriptsize{$\pm$12.93}$^*$&82.72\scriptsize{$\pm$18.13}       \\
(d)  & 4E\&D  & 4E\&D        & -            & 45.70\scriptsize{$\pm$27.79}&55.22\scriptsize{$\pm$28.73}&\underline{87.24}\scriptsize{$\pm$7.49}&72.01\scriptsize{$\pm$15.51}&62.69\scriptsize{$\pm$30.70}&68.05\scriptsize{$\pm$22.52}&89.13\scriptsize{$\pm$6.35}&67.91\scriptsize{$\pm$27.95}&68.49\scriptsize{$\pm$13.78}&83.00\scriptsize{$\pm$15.85}         \\
(e)  & 4E\&D  & 4E\&$\text{D}_{par}$       & -            & 47.67\scriptsize{$\pm$28.41}&56.35\scriptsize{$\pm$27.82}&\textbf{88.57}\scriptsize{$\pm$4.06}&74.22\scriptsize{$\pm$15.49}&{69.56}\scriptsize{$\pm$28.00}&\underline{71.56}\scriptsize{$\pm$15.90}&90.13\scriptsize{$\pm$5.66}&67.76\scriptsize{$\pm$27.65}$^*$&70.73\scriptsize{$\pm$13.49}&\textbf{83.83}\scriptsize{$\pm$15.14}     \\
(f)  & 4E\&D  & 4E\&$\text{D}_{par}$       & Mono-anchor  & \underline{48.32}\scriptsize{$\pm$31.53}&\underline{61.34}\scriptsize{$\pm$27.33}&85.19\scriptsize{$\pm$11.33}&\textbf{75.48}\scriptsize{$\pm$14.15}&67.37\scriptsize{$\pm$31.41}&70.52\scriptsize{$\pm$25.21}&\underline{90.67}\scriptsize{$\pm$5.43}&\underline{70.64}\scriptsize{$\pm$25.89}&\underline{71.19}\scriptsize{$\pm$12.40}&\underline{83.71}\scriptsize{$\pm$15.89}          \\ 
\hline
\rowcolor[HTML]{EFEFEF}(h) Ours     & 4E\&D  & 4E\&$\text{D}_{par}$        & Multi-anchor & 
\textbf{51.04}\scriptsize{$\pm$29.58}&\textbf{62.81}\scriptsize{$\pm$27.02}&84.76\scriptsize{$\pm$10.21}&\underline{74.27}\scriptsize{$\pm$13.20}&\textbf{72.84}\scriptsize{$\pm$26.11}&\textbf{73.87}\scriptsize{$\pm$20.79}&\textbf{91.02}\scriptsize{$\pm$5.28}&\textbf{72.08}\scriptsize{$\pm$24.90}&\textbf{72.84}\scriptsize{$\pm$11.47}&\textbf{83.83}\scriptsize{$\pm$15.43}\\ \hline
(g) \cite{dai2024federated}  & 4E\&D  & 4E       & Multi-anchor  & 46.56\scriptsize{$\pm$37.32}&56.37\scriptsize{$\pm$29.42}&85.66\scriptsize{$\pm$9.89}&73.26\scriptsize{$\pm$22.65}&{68.86}\scriptsize{$\pm$27.97}&71.67\scriptsize{$\pm$20.35}&89.97\scriptsize{$\pm$5.81}&69.88\scriptsize{$\pm$26.72}&70.27\scriptsize{$\pm$14.13}$^*$&83.23\scriptsize{$\pm$15.62}        \\ \hline
\end{tabular}
\end{adjustbox}
\label{tabs:ablation study}
\end{table*}

%% file: sections/tabs/tab7.tex
\begin{table*}[!t]\color{red}
\centering
\caption{Experimental results with different numbers of clients (four or six) and varying degrees of data heterogeneity (two or three modalities per client) on BraTS 2018 in mDSC (\%).
*: $p<0.05$ comparing against our method in each column.
Note that for ease of comparison, (1) the server and the clients in the same column use data of the same subset of subjects across settings; and (2) the Avg column here only averages the performance of the four clients that are consistent across settings, instead of all clients.
}
\setlength{\tabcolsep}{0.8mm}
\begin{adjustbox}{width=\textwidth}
\begin{tabular}{c|cccccccc}
\hline
\multicolumn{9}{c}{\textit{Two modalities per client $\times$ four clients}}\\
\hline
Method 
                        & F/t1c                & t1c/t2                  & F/t2                   & t1c/t1 &\phantom{55.96}-\phantom{\scriptsize{$\pm$21.7}3}&\phantom{55.96}-\phantom{\scriptsize{$\pm$21.7}3}                  & Avg                  & S                                        \\ \hline
Local models            & 78.57\scriptsize{$\pm$16.53}$^*$&76.27\scriptsize{$\pm$18.06}$^*$&49.02\scriptsize{$\pm$30.55}$^*$&55.96\scriptsize{$\pm$21.73}$^*$&\phantom{55.96}-\phantom{\scriptsize{$\pm$21.7}3}&\phantom{55.96}-\phantom{\scriptsize{$\pm$21.7}3}&64.96\scriptsize{$\pm$14.76}$^*$&79.18\scriptsize{$\pm$15.45}$^*$ \\
FedAvg \citep{mcmahan2017communication}                  & 66.98\scriptsize{$\pm$19.67}$^*$&71.09\scriptsize{$\pm$19.54}$^*$&50.73\scriptsize{$\pm$27.85}$^*$&42.22\scriptsize{$\pm$25.07}$^*$&\phantom{55.96}-\phantom{\scriptsize{$\pm$21.7}3}&\phantom{55.96}-\phantom{\scriptsize{$\pm$21.7}3}&57.76\scriptsize{$\pm$12.17}$^*$&76.54\scriptsize{$\pm$21.77}$^*$ \\
perFL \citep{wang2019federated}                   & 69.05\scriptsize{$\pm$20.31}$^*$&71.38\scriptsize{$\pm$18.98}$^*$&49.58\scriptsize{$\pm$28.43}$^*$&43.16\scriptsize{$\pm$24.91}$^*$&\phantom{55.96}-\phantom{\scriptsize{$\pm$21.7}3}&\phantom{55.96}-\phantom{\scriptsize{$\pm$21.7}3}&58.29\scriptsize{$\pm$12.17}$^*$&77.98\scriptsize{$\pm$23.26}$^*$ \\
FedNorm \citep{bernecker2022fednorm}                &  67.83\scriptsize{$\pm$20.85}$^*$&72.41\scriptsize{$\pm$16.84}$^*$&51.14\scriptsize{$\pm$27.88}$^*$&42.16\scriptsize{$\pm$25.18}$^*$&\phantom{55.96}-\phantom{\scriptsize{$\pm$21.7}3}&\phantom{55.96}-\phantom{\scriptsize{$\pm$21.7}3}&58.39\scriptsize{$\pm$12.26}$^*$&74.11\scriptsize{$\pm$19.06}$^*$\\
IOP-FL \citep{jiang2023iop}& 72.65\scriptsize{$\pm$16.97}$^*$&75.73\scriptsize{$\pm$18.23}$^*$&52.44\scriptsize{$\pm$23.46}$^*$&54.31\scriptsize{$\pm$19.66}$^*$&\phantom{55.96}-\phantom{\scriptsize{$\pm$21.7}3}&\phantom{55.96}-\phantom{\scriptsize{$\pm$21.7}3}&63.78\scriptsize{$\pm$12.54}$^*$&79.07\scriptsize{$\pm$15.76}$^*$  \\
FedCostWAvg \citep{machler2021fedcostwavg}& 73.80\scriptsize{$\pm$17.68}$^*$&74.16\scriptsize{$\pm$17.96}$^*$&55.86\scriptsize{$\pm$26.19}$^*$&54.52\scriptsize{$\pm$24.82}$^*$&\phantom{55.96}-\phantom{\scriptsize{$\pm$21.7}3}&\phantom{55.96}-\phantom{\scriptsize{$\pm$21.7}3}&63.58\scriptsize{$\pm$12.54}$^*$&79.11\scriptsize{$\pm$16.18}$^*$\\
FedPIDAvg \citep{machler2022fedpidavg} & 76.60\scriptsize{$\pm$17.15}$^*$&73.55\scriptsize{$\pm$19.41}$^*$&55.28\scriptsize{$\pm$27.90}$^*$&55.30\scriptsize{$\pm$24.49}$^*$&\phantom{55.96}-\phantom{\scriptsize{$\pm$21.7}3}&\phantom{55.96}-\phantom{\scriptsize{$\pm$21.7}3}&65.18\scriptsize{$\pm$11.94}$^*$&80.24\scriptsize{$\pm$17.63}$^*$ \\

CreamFL \citep{yu2023multimodal}                  & 79.67\scriptsize{$\pm$15.23}&76.99\scriptsize{$\pm$16.03}$^*$&61.87\scriptsize{$\pm$22.54}$^*$&66.57\scriptsize{$\pm$18.68}$^*$&\phantom{55.96}-\phantom{\scriptsize{$\pm$21.7}3}&\phantom{55.96}-\phantom{\scriptsize{$\pm$21.7}3}&71.28\scriptsize{$\pm$8.23}$^*$&80.52\scriptsize{$\pm$13.78}$^*$                          \\
FedIoT \citep{zhao2022multimodal}                 & \underline{80.09}\scriptsize{$\pm$14.88}&\underline{78.16}\scriptsize{$\pm$14.26}&\underline{62.31}\scriptsize{$\pm$23.38}&\underline{68.78}\scriptsize{$\pm$17.95}$^*$&\phantom{55.96}-\phantom{\scriptsize{$\pm$21.7}3}&\phantom{55.96}-\phantom{\scriptsize{$\pm$21.7}3}&\underline{72.34}\scriptsize{$\pm$7.20}$^*$&\underline{81.44}\scriptsize{$\pm$14.74}     \\
FedMSplit \citep{chen2022fedmsplit}              & 79.09\scriptsize{$\pm$16.14}$^*$&76.20\scriptsize{$\pm$16.74}$^*$&60.76\scriptsize{$\pm$27.76}$^*$&67.66\scriptsize{$\pm$21.34}$^*$&\phantom{55.96}-\phantom{\scriptsize{$\pm$21.7}3}&\phantom{55.96}-\phantom{\scriptsize{$\pm$21.7}3}&70.93\scriptsize{$\pm$9.83}$^*$&80.08\scriptsize{$\pm$16.27}$^*$                    \\ 
\hline
Ours   & \textbf{81.07}\scriptsize{$\pm$15.39}&\textbf{79.26}\scriptsize{$\pm$15.79}&\textbf{63.50}\scriptsize{$\pm$23.16}&\textbf{70.75}\scriptsize{$\pm$20.25}&\phantom{55.96}-\phantom{\scriptsize{$\pm$21.7}3}&\phantom{55.96}-\phantom{\scriptsize{$\pm$21.7}3}&\textbf{73.65}\scriptsize{$\pm$9.10}&\textbf{82.18}\scriptsize{$\pm$14.84} \\ \hline

\multicolumn{9}{c}{\textit{Two modalities per client $\times$ six clients}}
\end{tabular}
\end{adjustbox}

\setlength{\tabcolsep}{0.8mm}
\begin{adjustbox}{width=\textwidth}
\begin{tabular}{c|cccccccc}
\hline
{Method}                         & F/t1c                                  & t1c/t2                   & F/t2                  & t1c/t1              & t1/t2                  & F/t1          & Avg                & S               \\ \hline
Local models    &        78.57\scriptsize{$\pm$16.53}$^*$&76.27\scriptsize{$\pm$18.06}$^*$&49.02\scriptsize{$\pm$30.55}$^*$&55.96\scriptsize{$\pm$21.73}$^*$&44.54\scriptsize{$\pm$31.77}$^*$&41.11\scriptsize{$\pm$35.55}$^*$&64.96\scriptsize{$\pm$14.76}$^*$&79.18\scriptsize{$\pm$15.45}$^*$

    \\
FedAvg \citep{mcmahan2017communication}                  & 73.37\scriptsize{$\pm$17.60}$^*$&71.42\scriptsize{$\pm$20.66}$^*$&48.22\scriptsize{$\pm$29.24}$^*$&38.72\scriptsize{$\pm$26.88}$^*$&37.68\scriptsize{$\pm$32.95}$^*$&40.05\scriptsize{$\pm$34.43}$^*$&57.93\scriptsize{$\pm$15.12}$^*$&72.05\scriptsize{$\pm$19.50}$^*$
\\
perFL \citep{wang2019federated}                   & 75.91\scriptsize{$\pm$17.85}$^*$&71.14\scriptsize{$\pm$30.67}$^*$&55.86\scriptsize{$\pm$24.56}&37.50\scriptsize{$\pm$21.34}$^*$&41.05\scriptsize{$\pm$34.06}$^*$&39.60\scriptsize{$\pm$22.80}$^*$&60.10\scriptsize{$\pm$15.19}$^*$&74.10\scriptsize{$\pm$18.86}$^*$ \\
FedNorm \citep{bernecker2022fednorm}                & 76.61\scriptsize{$\pm$17.94}$^*$&71.34\scriptsize{$\pm$20.96}$^*$&47.11\scriptsize{$\pm$26.98}$^*$&46.12\scriptsize{$\pm$22.27}$^*$&34.65\scriptsize{$\pm$29.10}$^*$&38.16\scriptsize{$\pm$30.31}$^*$&60.30\scriptsize{$\pm$15.97}$^*$&67.37\scriptsize{$\pm$22.51}$^*$               \\
IOP-FL \citep{jiang2023iop}& 76.61\scriptsize{$\pm$16.79}$^*$  & 75.36\scriptsize{$\pm$19.22}$^*$ & 56.61\scriptsize{$\pm$25.57}$^*$ & 59.35\scriptsize{$\pm$19.37}$^*$ & 43.12\scriptsize{$\pm$28.37}$^*$ & 42.91\scriptsize{$\pm$32.42}$^*$& 66.98\scriptsize{$\pm$13.30}$^*$ & 79.58\scriptsize{$\pm$17.47}$^*$ \\
FedCostWAvg \citep{machler2021fedcostwavg}& 76.48\scriptsize{$\pm$16.75}$^*$&76.06\scriptsize{$\pm$15.27}$^*$&57.22\scriptsize{$\pm$24.35}$^*$&60.74\scriptsize{$\pm$16.76}$^*$&45.35\scriptsize{$\pm$29.00}$^*$&41.85\scriptsize{$\pm$35.91}$^*$&67.63\scriptsize{$\pm$12.67}$^*$&{81.74}\scriptsize{$\pm$15.16}$^*$\\
FedPIDAvg \citep{machler2022fedpidavg}& 76.51\scriptsize{$\pm$15.46}$^*$&74.79\scriptsize{$\pm$15.45}$^*$&57.35\scriptsize{$\pm$24.24}$^*$&64.21\scriptsize{$\pm$18.05}$^*$&46.58\scriptsize{$\pm$27.30}$^*$&44.84\scriptsize{$\pm$32.37}$^*$&68.22\scriptsize{$\pm$11.71}$^*$&\underline{82.18}\scriptsize{$\pm$14.72}$^*$\\
CreamFL \citep{yu2023multimodal}                  & 78.31\scriptsize{$\pm$16.54}&73.44\scriptsize{$\pm$15.79}$^*$&60.00\scriptsize{$\pm$26.30}$^*$&65.01\scriptsize{$\pm$20.44}$^*$&52.51\scriptsize{$\pm$27.17}&44.25\scriptsize{$\pm$33.79}$^*$&69.19\scriptsize{$\pm$17.68}$^*$&79.80\scriptsize{$\pm$15.50}$^*$           \\ 
FedIoT \citep{zhao2022multimodal}                 & 79.66\scriptsize{$\pm$15.27}&\underline{79.92}\scriptsize{$\pm$13.16}&{62.85}\scriptsize{$\pm$19.32}&\underline{70.51}\scriptsize{$\pm$25.25}&\textbf{53.89}\scriptsize{$\pm$17.34}&\underline{46.24}\scriptsize{$\pm$31.23}$^*$&\underline{73.24}\scriptsize{$\pm$18.00}&81.51\scriptsize{$\pm$20.20}$^*$                    \\ 
FedMSplit \citep{chen2022fedmsplit}               & \underline{79.87}\scriptsize{$\pm$16.17}&77.61\scriptsize{$\pm$14.14}&\underline{62.88}\scriptsize{$\pm$25.31}&65.35\scriptsize{$\pm$22.20}$^*$&46.98\scriptsize{$\pm$28.60}$^*$&44.64\scriptsize{$\pm$35.23}$^*$&71.41\scriptsize{$\pm$13.53}$^*$&78.75\scriptsize{$\pm$15.92}$^*$                       \\
 \hline

Ours   & \textbf{81.40}\scriptsize{$\pm$15.27}&\textbf{80.13}\scriptsize{$\pm$14.94}&\textbf{63.22}\scriptsize{$\pm$24.63}&\textbf{72.76}\scriptsize{$\pm$19.31}&\underline{53.49}\scriptsize{$\pm$28.23}&\textbf{49.51}\scriptsize{$\pm$34.32}&\textbf{74.13}\scriptsize{$\pm$12.35}&\textbf{82.89}\scriptsize{$\pm$13.84} \\ \hline
\end{tabular}
\end{adjustbox}

\setlength{\tabcolsep}{0.8mm}
\begin{adjustbox}{width=\textwidth}
\begin{tabular}{c|cccccccc}
\multicolumn{9}{c}{\textit{Three modalities per client $\times$ four clients}} \\
\hline
Method           & F/t1c/t1                & F/t1c/t2        & F/t1/t2            & t1c/t1/t2        &\phantom{55.96}-\phantom{\scriptsize{$\pm$21.7}3}&\phantom{55.96}-\phantom{\scriptsize{$\pm$21.7}3}            & Avg                  & S  \\   \hline                
Local models            & 77.90\scriptsize{$\pm$18.16}$^*$&76.20\scriptsize{$\pm$17.85}$^*$&43.95\scriptsize{$\pm$29.32}$^*$&62.12\scriptsize{$\pm$25.07}$^*$&\phantom{55.96}-\phantom{\scriptsize{$\pm$21.7}3}&\phantom{55.96}-\phantom{\scriptsize{$\pm$21.7}3}&65.04\scriptsize{$\pm$13.04}$^*$&79.18\scriptsize{$\pm$15.45}$^*$\\
FedAvg \citep{mcmahan2017communication}                  & 76.01\scriptsize{$\pm$18.55}$^*$&75.22\scriptsize{$\pm$18.10}$^*$&42.45\scriptsize{$\pm$28.73}$^*$&61.85\scriptsize{$\pm$22.31}$^*$&\phantom{55.96}-\phantom{\scriptsize{$\pm$21.7}3}&\phantom{55.96}-\phantom{\scriptsize{$\pm$21.7}3}&63.88\scriptsize{$\pm$14.76}$^*$&75.79\scriptsize{$\pm$19.65}$^*$\\
perFL \citep{wang2019federated}                   & 77.24\scriptsize{$\pm$17.43}$^*$&78.74\scriptsize{$\pm$16.91}$^*$&41.99\scriptsize{$\pm$29.14}$^*$&63.24\scriptsize{$\pm$21.12}$^*$&\phantom{55.96}-\phantom{\scriptsize{$\pm$21.7}3}&\phantom{55.96}-\phantom{\scriptsize{$\pm$21.7}3}&65.30\scriptsize{$\pm$14.76}$^*$&78.26\scriptsize{$\pm$17.68}$^*$\\
FedNorm \citep{bernecker2022fednorm}               & 69.92\scriptsize{$\pm$19.75}$^*$&78.55\scriptsize{$\pm$16.74}$^*$&48.63\scriptsize{$\pm$28.08}$^*$&58.96\scriptsize{$\pm$23.43}$^*$&\phantom{55.96}-\phantom{\scriptsize{$\pm$21.7}3}&\phantom{55.96}-\phantom{\scriptsize{$\pm$21.7}3}&64.02\scriptsize{$\pm$11.27}$^*$&73.31\scriptsize{$\pm$19.51}$^*$ \\
IOP-FL \citep{jiang2023iop}& 76.55\scriptsize{$\pm$17.41}$^*$&77.32\scriptsize{$\pm$18.05}$^*$&41.82\scriptsize{$\pm$28.90}$^*$&67.27\scriptsize{$\pm$18.07}$^*$&\phantom{55.96}-\phantom{\scriptsize{$\pm$21.7}3}&\phantom{55.96}-\phantom{\scriptsize{$\pm$21.7}3}&66.27\scriptsize{$\pm$15.31}$^*$&76.43\scriptsize{$\pm$17.62}$^*$ \\ 
FedCostWAvg \citep{machler2021fedcostwavg}& 77.17\scriptsize{$\pm$16.40}$^*$&79.09\scriptsize{$\pm$16.47}$^*$&40.76\scriptsize{$\pm$29.56}$^*$&68.06\scriptsize{$\pm$19.02}$^*$&\phantom{55.96}-\phantom{\scriptsize{$\pm$21.7}3}&\phantom{55.96}-\phantom{\scriptsize{$\pm$21.7}3}&66.27\scriptsize{$\pm$15.31}$^*$&76.60\scriptsize{$\pm$16.96}$^*$\\
FedPIDAvg \citep{machler2022fedpidavg}& 77.03\scriptsize{$\pm$17.02}$^*$&79.48\scriptsize{$\pm$16.17}$^*$&38.04\scriptsize{$\pm$30.56}$^*$&64.92\scriptsize{$\pm$23.91}$^*$&\phantom{55.96}-\phantom{\scriptsize{$\pm$21.7}3}&\phantom{55.96}-\phantom{\scriptsize{$\pm$21.7}3}&64.87\scriptsize{$\pm$16.44}$^*$&76.25\scriptsize{$\pm$18.02}$^*$\\
CreamFL \citep{yu2023multimodal}                   & 77.96\scriptsize{$\pm$16.44}&79.95\scriptsize{$\pm$15.68}&58.00\scriptsize{$\pm$26.49}$^*$&\underline{74.34}\scriptsize{$\pm$17.52}&\phantom{55.96}-\phantom{\scriptsize{$\pm$21.7}3}&\phantom{55.96}-\phantom{\scriptsize{$\pm$21.7}3}&72.56\scriptsize{$\pm$8.65}$^*$&80.36\scriptsize{$\pm$15.36}$^*$                          \\ 
FedIoT \citep{zhao2022multimodal}                & \underline{79.86}\scriptsize{$\pm$15.49}&\underline{81.45}\scriptsize{$\pm$14.11}&\underline{60.78}\scriptsize{$\pm$24.99}$^*$&73.95\scriptsize{$\pm$14.67}&\phantom{55.96}-\phantom{\scriptsize{$\pm$21.7}3}&\phantom{55.96}-\phantom{\scriptsize{$\pm$21.7}3}&\underline{74.01}\scriptsize{$\pm$8.21}$^*$&\underline{82.05}\scriptsize{$\pm$14.38}$^*$                           \\ 
FedMSplit \citep{chen2022fedmsplit}               & 77.26\scriptsize{$\pm$16.95}$^*$&80.54\scriptsize{$\pm$15.12}&59.88\scriptsize{$\pm$26.56}$^*$&71.61\scriptsize{$\pm$17.71}&\phantom{59.96}-\phantom{\scriptsize{$\pm$21.7}3}&\phantom{55.96}-\phantom{\scriptsize{$\pm$21.7}3}&72.32\scriptsize{$\pm$9.47}$^*$&78.49\scriptsize{$\pm$16.55}$^*$                                 \\\hline
Ours & \textbf{81.51}\scriptsize{$\pm$14.67}&\textbf{81.78}\scriptsize{$\pm$15.42}&\textbf{62.01}\scriptsize{$\pm$23.22}&\textbf{74.54}\scriptsize{$\pm$13.98}&\phantom{55.96}-\phantom{\scriptsize{$\pm$21.7}3}&\phantom{55.96}-\phantom{\scriptsize{$\pm$21.7}3}&\textbf{74.96}\scriptsize{$\pm$8.55}&\textbf{82.35}\scriptsize{$\pm$12.86}\\ \hline
\end{tabular}
\end{adjustbox}
\label{tab:varying_num}

\end{table*}

%% file: sections/4-discussion-conclusion.tex
\section{Discussion}

\subsection{Deep Analysis of FedMEPD Framework}


Unlike existing federated learning (FL) methods for medical images that mostly considered intramodal heterogeneity, this work brought forward and aimed to address the intermodal heterogeneity issue in multimodal FL.
Specifically, some FL participants may possess only a subset of the complete imaging modalities, posing intermodal heterogeneity as a challenge to effectively training a global model on all participants' data.
Meanwhile, each participant also expects an optimal model tailored to its local data characteristics (\textit{e.g.}, missing certain modalities) by participating in the FL.
To address the intermodal heterogeneity and the concurrent targets of optimal global and local models, this work proposed a novel FL framework with \textbf{fed}erated \textbf{m}odality-specific \textbf{e}ncoders and \textbf{p}artially \textbf{p}ersonalized multimodal fusion \textbf{d}ecoders (FedMEPD).
The central motivation of FedMEPD is to simultaneously facilitate global sharing of common knowledge and adaptive personalization of local models within the FL paradigm.
Thus, FedMEPD yielded not only an optimal global model that worked well with full-modal data but also optimal personalized models that worked well in specific missing-modal situations for the clients.
The experimental results on two benchmark datasets (Table \ref{tabs:state-of-art}) showed that FedMEPD outperformed a series of classical and recent works focused on either personalized FL or multimodal FL, validating our motivation.

%
%

Above all, FedMEPD employed a late-fusion multimodal network architecture \citep{ding2021rfnet} as backbone.
The modality-specific encoders allowed flexible handling of the intermodal heterogeneity across FL participants in the first place, by aggregating the encoder parameters for a specific modality only from those clients with data of that modality.
The comparison between rows (a) and (c) in Table \ref{tabs:ablation study} showed that employing and federating an exclusive encoder for each modality significantly improved the clients' average performance (13.33\% absolute increase in mDSC) upon the baseline, validating our design of federated modality-specific encoders.

Meanwhile, FedMEPD adopted a partially federating, partially personalizing strategy for the multimodal fusion decoders.
The strategy was guided by the consistency between the parameter updates to the server and clients: only those filters with consistent updates between them were federated, whereas others were preserved for personalization adapted to the clients' local data.
The superior performance of row (e) to rows (c) and (d) in Table \ref{tabs:ablation study} confirmed the strategy's efficacy compared with fully personalizing or fully federating the decoders.
We attribute these advantages to the concurrent common knowledge sharing and local model personalization enabled by the strategy.
A recent work, GRACE, similarly leveraged the gradient consistency across clients to enhance global gradients \citep{zhang2023grace}.
However, it did not personalize the client models based on global-client consistency like FedMEPD.



\begin{figure}[t]
\centering
\includegraphics[trim=0 24 0 0, clip, width=.8\textwidth]{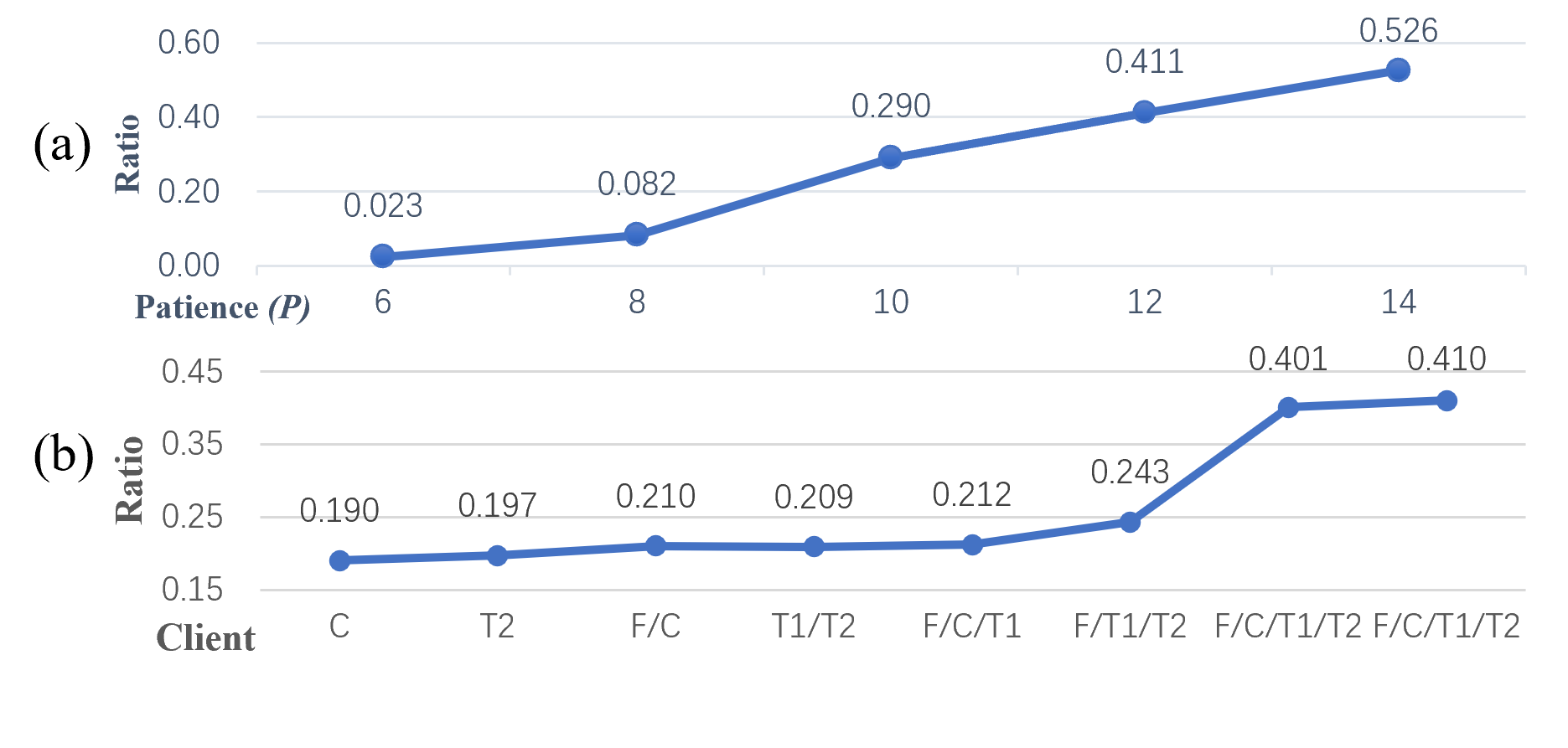} 
\caption{\color{blue}(a) Total ratio of federated decoder parameters of all clients concerning different personalization patience $P$. 
(b) Ratio of federated decoder parameters for each client with $P=10$.
}
\label{fig:ratio}
\end{figure}

To gain deeper insights into the proposed partially federating, partially personalizing strategy, we visualize the total ratio of federated decoder parameters of all clients with varying $P$ values (the personalization patience hyper-parameter in Eqn. (\ref{eq:fed_or_per})) in Fig. \ref{fig:ratio}(a).
The ratio increases as $P$ increases, indicating fewer parameters are personalized.
This trend validates our design of the regulating hyper-parameter.
When $P=10$, the total ratio of federated parameters is 0.29;
in other words, 71\% parameters are personalized.
Fig. \ref{fig:ratio}(b) further visualizes the ratio of federated decoder parameters of each client with $P=10$.
The ratio increases as data modalities increase.
This is reasonable, as more modalities present smaller modality gaps compared with the server-end full-modal data.
Therefore, our strategy adaptively adjusts the appropriate extent of personalization for clients with different modality combinations.
{\color{red}Particularly, the ratios for the two clients with full-modal data are both around 0.4, instead of close to 1. 
We conjecture that this phenomenon results from the cross-site intramodal heterogeneity;
that is, although the two clients and the server have the same modalities, domain gaps may still exist due to institutional variations. 
By design, our strategy can also handle such intramodal heterogeneity, in addition to the intermodal heterogeneity.
Therefore, the federated ratios of 0.4 actually indicate the intramodal domain gaps between the two clients and the server. 
We have further experimented with applying the server decoder parameters to the two full-modal clients to make the federated ratio 100\%. 
On BraTS 2018, mDSCs of these two clients drop from 90.69 and 69.62 to 88.37 and 49.91, respectively, indicating that the ratios around 0.4 are more appropriate.
These results confirm the efficacy of our strategy in handling cross-site intramodal heterogeneity.}

Lastly, FedMEPD proposed adaptively calibrating local clients' missing-modal representation towards the multi-anchor full-modal prototypes extracted by the server.
The calibration process recovered part of the absent information due to missing modalities via scaled cross-attention operations between the local representations and the global prototypes.
In Table \ref{tabs:ablation study}, row (f) showed that multimodal representations were effective in calibrating local missing-modal features; row (h) further demonstrated the superior efficacy of multiple anchors in representing 3D multimodal medical images to mono ones.

{\color{red}Besides being effective, the adaptive multi-anchor calibration is also computationally efficient.
On the one hand, the anchors are extracted at the server via K-means clustering.
Concretely, we use image-wise per-class features obtained by masked average pooling at the most abstract (smallest) feature scale for clustering.
For the amounts of data used in our experiments ($\sim$50--80 images), it was done in 0.1 seconds.
We have further experimented with clustering 10,000 toy features of the same dimension to simulate the scenarios of significantly more server data, which only takes about 1 seconds.
On the other hand, the adaptive calibration at each client involves scaled dot-product cross-attention.
With the backbone network (and thus the feature dimensions) fixed, the computational complexity mainly depends on $N_k$, the number of anchors for each segmentation class.
With the optimal $N_k=4$ determined in Table \ref{tabs:clusNum}, the GPU memories required for training and inference of a full-modal client (which demands the most resources) are around 11.2 and 3.7 GB, respectively.
These requirements can be easily met by the hardware used in the past few years.
In addition, one RTX 2080 Ti GPU can infer three full-modal images in 10 seconds.
Therefore, our FedMEPD framework is highly scalable and only requires reasonable resources for training and inference.
}



{\color{red}
\subsection{Insights into Experiments and Results}
For the main experiments on the BraTS datasets, the federated data split largely followed our practical problem setting, while also being restricted by the datasets used for experiments.
On the one hand, as stated in the problem definition (Section 3.1), we assumeed the server to be a major regional hospital (e.g., a large, comprehensive hospital in a big city), which naturally housed significantly more data of full modalities than the clients that were assumed to be local health units (e.g., rural clinics). 
On the other hand, our experiments run a strict federated learning (FL) simulation by dividing data from different centers into separate clients (per the datasets' official information). 
In addition, the local training, validation, and testing split per site was appropriate for evaluating personalized FL, where each participant expected a personalized model tailored to its local data characteristics, and thus should be evaluated with its own test data. 
Therefore, our setup genuinely reflected real-world scenarios of hetero-modal medical imaging FL. 
However, the scales and inherent data distributions of the BraTS 2018 and 2020 datasets inevitably constrained the clients to possess only 20--35 subjects under this setting. 
The local 6:2:2 data split to ensure routine training, validation, and testing data isolation for standard machine learning further resulted in low numbers of testing subjects per client. 
Hence, the small number of testing samples of each client might have resulted in the absence of statistical significance in many comparisons for BraTS 2018, e.g., in Table \ref{tabs:state-of-art} and Table \ref{tabs:ablation study}.

Additionally, in Table \ref{tabs:state-of-art}, the baseline local models were outperformed likely because they overfitted on the small number of training samples on their respective clients. 
Nonetheless, this challenge is precisely one of the motivations of our work: to obtain better personalized models in the case of limited local data with incomplete modalities through FL. 
Therefore, we consider the superior performance of our method to the local models as supporting evidence of our method's effectiveness in heterogeneous multimodal medical imaging FL. 
Meanwhile, the server had 88 and 129 subjects of full-modal data for the BraTS 2018 and 2020 datasets, respectively. 
Even with such relatively sufficient data for local training of segmentation models, our method still demonstrated substantial performance improvements upon the already competitive local models regarding server performance (e.g., 84.98 versus 82.56 on BraTS 2018).
The superior server performance indicated that our framework effectively made use of the clients' hetero-modal data to improve the global model's full-modal performance, too.
In conclusion, our FedMEPD framework successfully fulfilled the dual motivations of this work: obtaining an optimal global model and personalized client models in hetero-modal medical imaging FL scenarios.
}

{\color{red}
\subsection{Generalizability to Other Medical Imaging Data and Tasks}

To evaluate the generalizability of our FedMEPD framework to other medical imaging data and tasks beyond brain tumor segmentation in multi-parameter MRI, we conduct additional experiments using the HaN-Seg dataset \citep{podobnik2023han}.
HaN-Seg consists of 42 paired computed tomography (CT) and T1-weighted MRI images that are publicly available, as well as manual delineations of 30 head and neck organ-at-risk (OARs).
The paired CT and MRI images were deformably registered following \cite{podobnik2023han}.
We select eight OARs that are consistently annotated across all subjects as our segmentation targets: mandible, buccal mucosa, oral cavity, submandibular gland (left), submandibular gland (right), lips, parotid gland (right), and spinal cord.
Of the 42 subjects, nine ($\sim$20\%) are used as the test set, whereas the remaining ones are evenly distributed to the server, a CT client, and an MRI client for training.
Table \ref{tab::sota_han} shows the results, including a comparison with other methods.
As we can see, the comparative trends among the methods resonate well with those on the BraTS datasets, with our method achieving the best performance for all computed metrics.
Notably, our CT and average client mDSCs are higher by absolute margins of 2.14\% and 1.49\% than the second-best numbers.
Also, our framework substantially improves upon the baseline local models by absolute margins of {13.97--16.72\%}.
These results demonstrate the effectiveness and generalizability of our framework in hetero-modal medical imaging FL across data and tasks.}


\input{sections/tabs/tab_sota_han}

\subsection{Limitations and Future work}
This work had limitations.
{\color{red}
Our experiments roughly balanced the number of samples at each client while ensuring all cases sourced from the same institute were assigned to the same client.
In practice, imbalanced client sizes---meaning that the amount of training data available at each client differs substantially---are another challenge to effective FL.
We are aware of strategies for dynamic weight assignment for parameter aggregation based on client data sizes \cite[e.g.,][]{hsu2020federated}. 
Although this work was focused on intermodal heterogeneity and minimized the impact of client imbalance in its experimental setup, we expect that incorporating such strategies would make our method more robust in future practical applications.
}

%

\section{Conclusion}
This work focused primarily on intermodal heterogeneity issues caused by participants with incomplete modalities in multimodal medical image federated learning (FL).
To address the issues, we proposed FedMEPD, a novel FL framework with federated modality-specific encoders and partially personalized multimodal fusion decoders.
Experiments on two public benchmarks showed that FedMEPD outperformed state-of-the-art works on personalized or multimodal FL in addressing the issues above, and its novel designs were effective.

%% file: sections/tabs/tab_sota_han.tex
\begin{table*}[t]\color{red}
\centering
\setlength{\tabcolsep}{0.8mm}
\caption{Experimental results on the HaN-Seg dataset \citep{podobnik2023han} in mDSC (\%).
CT and MRI indicate the performance of the clients with the corresponding modalities, Average indicates their average, and Full indicates the server's performance.
*: $p<0.05$ comparing against our method in each column.
}%
\begin{adjustbox}{width=0.8\textwidth}
\begin{tabular}{c|cccc}
\hline
Method &    CT & MRI                & Average                  & Full                           \\ \hline
Local models            & 54.10\scriptsize{$\pm$2.69}$^*$&51.37\scriptsize{$\pm$2.69}$^*$&52.74\scriptsize{$\pm$2.36}$^*$&61.58\scriptsize{$\pm$3.88}$^*$      \\ 
RFNet \citep{ding2021rfnet} & 66.57\scriptsize{$\pm$4.42}$^*$ &64.32\scriptsize{$\pm$4.91}$^*$& 65.45\scriptsize{$\pm$2.58}$^*$ & 74.13\scriptsize{$\pm$3.79}$^*$ \\
\hline
FedAvg \citep{mcmahan2017communication}          &  51.60\scriptsize{$\pm$2.90}$^*$&49.11\scriptsize{$\pm$2.90}$^*$&50.35\scriptsize{$\pm$1.25}$^*$&55.80\scriptsize{$\pm$3.65}$^*$      \\
perFL \citep{wang2019federated}                   & 51.81\scriptsize{$\pm$2.04}$^*$&50.79\scriptsize{$\pm$2.84}$^*$&50.80\scriptsize{$\pm$2.01}$^*$&56.69\scriptsize{$\pm$2.98}$^*$    \\
FedNorm \citep{bernecker2022fednorm}                & 49.67\scriptsize{$\pm$3.22}$^*$&47.99\scriptsize{$\pm$2.95}$^*$&48.83\scriptsize{$\pm$1.98}$^*$&54.30\scriptsize{$\pm$3.45}$^*$       \\
IOP-FL \citep{jiang2023iop}          &53.81\scriptsize{$\pm$2.49}$^*$&52.66\scriptsize{$\pm$2.23}$^*$&53.24\scriptsize{$\pm$2.12}$^*$&57.02\scriptsize{$\pm$2.71}$^*$ \\  
FedCostWAvg \citep{machler2021fedcostwavg}&51.44\scriptsize{$\pm$1.63}$^*$&48.42\scriptsize{$\pm$3.50}$^*$&49.93\scriptsize{$\pm$1.51}$^*$&55.95\scriptsize{$\pm$4.31}$^*$  \\
FedPIDAvg \citep{machler2022fedpidavg}&52.25\scriptsize{$\pm$1.76}$^*$&49.82\scriptsize{$\pm$2.79}$^*$&51.04\scriptsize{$\pm$1.99}$^*$&56.71\scriptsize{$\pm$3.68}$^*$ \\
CreamFL \citep{yu2023multimodal}                  & 61.58\scriptsize{$\pm$4.20}&59.90\scriptsize{$\pm$3.87}&60.74\scriptsize{$\pm$3.01}$^*$&71.13\scriptsize{$\pm$3.17}$^*$  \\ 
FedIoT \citep{zhao2022multimodal}                 & \underline{68.68}\scriptsize{$\pm$5.18}&64.50\scriptsize{$\pm$2.67}&\underline{66.59}\scriptsize{$\pm$2.09}&76.05\scriptsize{$\pm$2.66}     \\ 
FedMSplit \citep{chen2022fedmsplit}              &  67.79\scriptsize{$\pm$5.24}&\underline{65.01}\scriptsize{$\pm$3.03}&66.40\scriptsize{$\pm$1.96}&\underline{76.11}\scriptsize{$\pm$2.52}     \\ 
 \hline
Ours   & \textbf{70.82}\scriptsize{$\pm$4.62}&\textbf{65.34}\scriptsize{$\pm$4.42}&\textbf{68.08}\scriptsize{$\pm$2.74}&\textbf{76.59}\scriptsize{$\pm$2.77}
 \\ 
\hline
\end{tabular}
\end{adjustbox}
\label{tab::sota_han}

\end{table*}